\documentclass[acmtog,nonacm,anonymous=False,review=False]{acmart}
\acmSubmissionID{PAPERS\_666}

\usepackage{booktabs} % For formal tables
\usepackage{appendix}
\renewcommand\footnotetextcopyrightpermission[1]{} % removes footnote with conference information in first column
\usepackage{cleveref}
\usepackage{colortbl}
\usepackage{graphicx} 
\usepackage[ruled]{algorithm2e} % For algorithms

\SetAlFnt{\small}
\SetAlCapFnt{\small}
\SetAlCapNameFnt{\small}
\SetAlCapHSkip{0pt}
\usepackage{multirow}
\usepackage{enumitem}
\usepackage{color} 
  
\usepackage{bbding}

\definecolor{applegreen}{rgb}{0.55, 0.71, 0.0}
\definecolor{autumnorange}{rgb}{0.87, 0.61, 0.33}
\definecolor{3rd}{HTML}{FCF596}
\definecolor{2nd}{HTML}{FBD288}
\definecolor{1st}{HTML}{FF9C73}

\definecolor{ours}{HTML}{E5E1DA}
\definecolor{removal}{HTML}{E1F0DA}
\definecolor{integrate}{HTML}{D1E9F6}
\definecolor{replace}{HTML}{FFECC8}

\newcommand{\modelname}{AnimPortrait3D\xspace}

\acmJournal{TOG}
\acmYear{2025}
\begin{document}
% Title portion
\title{Text-based Animatable 3D Avatars with Morphable Model Alignment}

\author{Yiqian Wu}
\affiliation{%
     \institution{ETH Zürich; State Key Lab of CAD\&CG, Zhejiang University}
     \city{Zürich}
     \country{Switzerland}
     }
\orcid{0000-0002-2432-809X}
\email{onethousand1250@gmail.com}

\author{Malte Prinzler}
\affiliation{%
     \institution{ETH Zürich}
     \city{Zürich}
     \country{Switzerland}
     }
\orcid{0009-0003-4268-9133}
\email{malte.prinzler@tuebingen.mpg.de}

\author{Xiaogang Jin}
\authornote{Corresponding author.}
\affiliation{%
     \institution{State Key Lab of CAD\&CG, Zhejiang University}
     \city{Hangzhou}
     \country{China}
     }
\orcid{0000-0001-7339-2920}
\email{jin@cad.zju.edu.cn}

\author{Siyu Tang} 
\affiliation{%
     \institution{ETH Zürich}
     \city{Zürich}
     \country{Switzerland} 
     }
\orcid{0000-0002-1015-4770}
\email{siyu.tang@inf.ethz.ch}

\begin{abstract} 
The generation of high-quality, animatable 3D head avatars from text has enormous potential in content creation applications such as games, movies, and embodied virtual assistants. Current text-to-3D generation methods typically combine parametric head models with 2D diffusion models using score distillation sampling to produce 3D-consistent results. However, they struggle to synthesize realistic details and suffer from misalignments between the appearance and the driving parametric model, resulting in unnatural animation results. We discovered that these limitations stem from ambiguities in the 2D diffusion predictions during 3D avatar distillation, specifically: i) the avatar's appearance and geometry is underconstrained by the text input, and ii) the semantic alignment between the predictions and the parametric head model is insufficient because the diffusion model alone cannot incorporate information from the parametric model. 
In this work, we propose a novel framework, \textit{\modelname}, for text-based realistic animatable 3DGS avatar generation with morphable model alignment, and introduce two key strategies to address these challenges. First, we tackle appearance and geometry ambiguities by utilizing prior information from a pretrained text-to-3D model to initialize a 3D avatar with robust appearance, geometry, and rigging relationships to the morphable model. Second, we refine the initial 3D avatar for dynamic expressions using a ControlNet that is conditioned on semantic and normal maps of the morphable model to ensure accurate alignment. As a result, our method outperforms existing approaches in terms of synthesis quality, alignment, and animation fidelity. Our experiments show that the proposed method advances the state of the art in text-based, animatable 3D head avatar generation.
\textbf{Code and model for this paper are at \href{https://github.com/oneThousand1000/AnimPortrait3D}{\color{cyan}{AnimPortrait3D}}.}

\end{abstract}

%
% The code below should be generated by the tool at
% p
% Please copy and paste the code instead of the example below.
%
\begin{CCSXML}
<ccs2012>
   <concept>
       <concept_id>10010147.10010178.10010224</concept_id>
       <concept_desc>Computing methodologies~Computer vision</concept_desc>
       <concept_significance>500</concept_significance>
       </concept>
 </ccs2012>
\end{CCSXML}

\ccsdesc[500]{Computing methodologies~Computer vision}
%
% End generated code
%

\keywords{Animatable 3D avatar generation, Gaussian splatting, diffusion models}

\begin{teaserfigure}
  \centering
  \includegraphics[width=0.99 \linewidth]{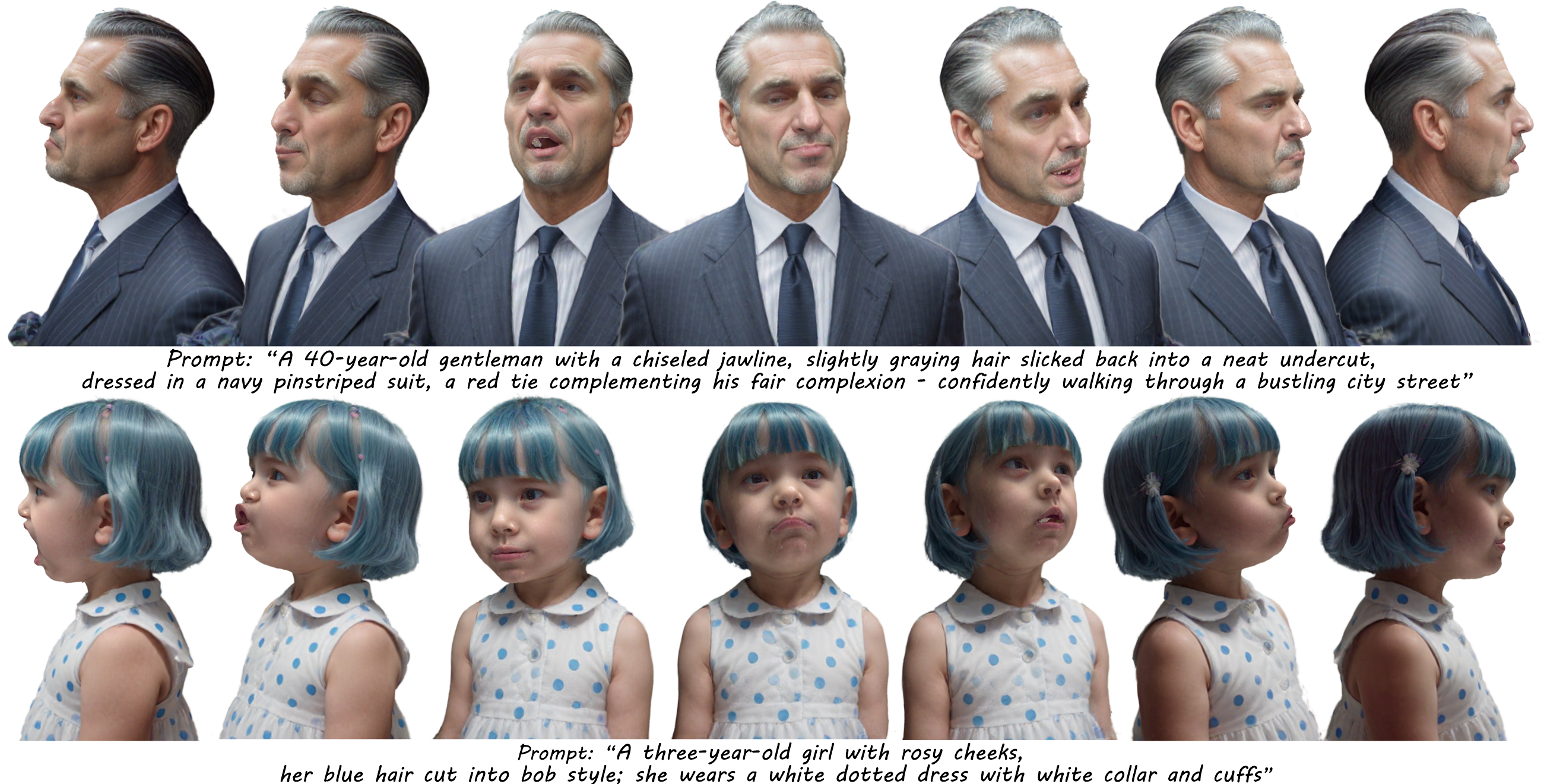}
  \caption{Our method generates high-quality, realistic, and animatable 3D avatars from text descriptions which can be driven with morphable model parameters.
  % Our method can generate high-quality, realistic, and animatable 3D avatars from text descriptions.
  % input. 
  We show the rendered results of our generated avatars from various camera angles, with expressions and body poses sampled from the NeRSemble dataset \cite{NeRSemble}.
  %We present rendered results of our generated avatars, displaying them from various camera angles and with random expressions and body poses. %\MP{looks nice! Maybe add quotation marks around text prompts and slightly more spacing between the images.}
  } 
\label{fig:teaser}
\end{teaserfigure}

 \maketitle

\section{Introduction}
\label{sec: Introduction}
The creation of 3D human avatars is an extensively researched topic with applications in gaming, filmmaking, and social media. 
%Reconstruction methods based on captured data can produce highly detailed and realistic results \cite{NPGA, GaussianAvatars, SplattingAvatar}, but they necessitate complex and laborious data collection processes. In contrast, feedforward networks trained on large amounts of 2D data \cite{LivePortrait,X-Portrait} can generate 2D video animations from text or images, but they frequently lack 3D consistency.
Text-driven methods have demonstrated impressive results in generating 3D avatars using 2D diffusion models \cite{joker, ProlificDreamer, magicmirror, AvatarStudio, MVDream, HumanNorm, DreamHuman}. Furthermore, parametric models \cite{smplx, flame, bfm} offer a viable approach for creating animatable avatars \cite{HeadStudio, TADA, HumanGaussian} for subsequent applications. 

However, existing diffusion-model-based generation methods primarily rely on algorithms such as score distillation sampling (SDS) \cite{DreamFusion} to integrate the 2D information from diffusion models into a 3D avatar, introducing ambiguities that reduce the quality of the results. 
These ambiguities stem from two major factors:  i) First, a single text prompt in a diffusion model maps to a wide range of images, resulting in underconstrained appearance and geometry guidance. This leads to blurred results and the ``Janus'' problem \cite{DreamFusion, ProlificDreamer}. Portrait3D \cite{portrait3d} addresses this by utilizing an appearance-geometry joint prior. 
However, the results exhibit noisy geometry and are restricted to static reconstructions without animation capabilities. 
ii) Second, the diffusion model's predictions typically align poorly with the underlying parametric model which causes artifacts during animation.  %
HeadStudio \cite{HeadStudio} aims to solve this by using a landmarks-based ControlNet for guidance, but the 3D information provided by landmarks alone remains insufficient for robust alignment. The primary challenge is to incorporate robust guidance that is aware of both geometry and semantic information from the parametric model into the optimization process. 
 
\begin{figure*}[t]
  \centering
  \includegraphics[width=0.99\linewidth]{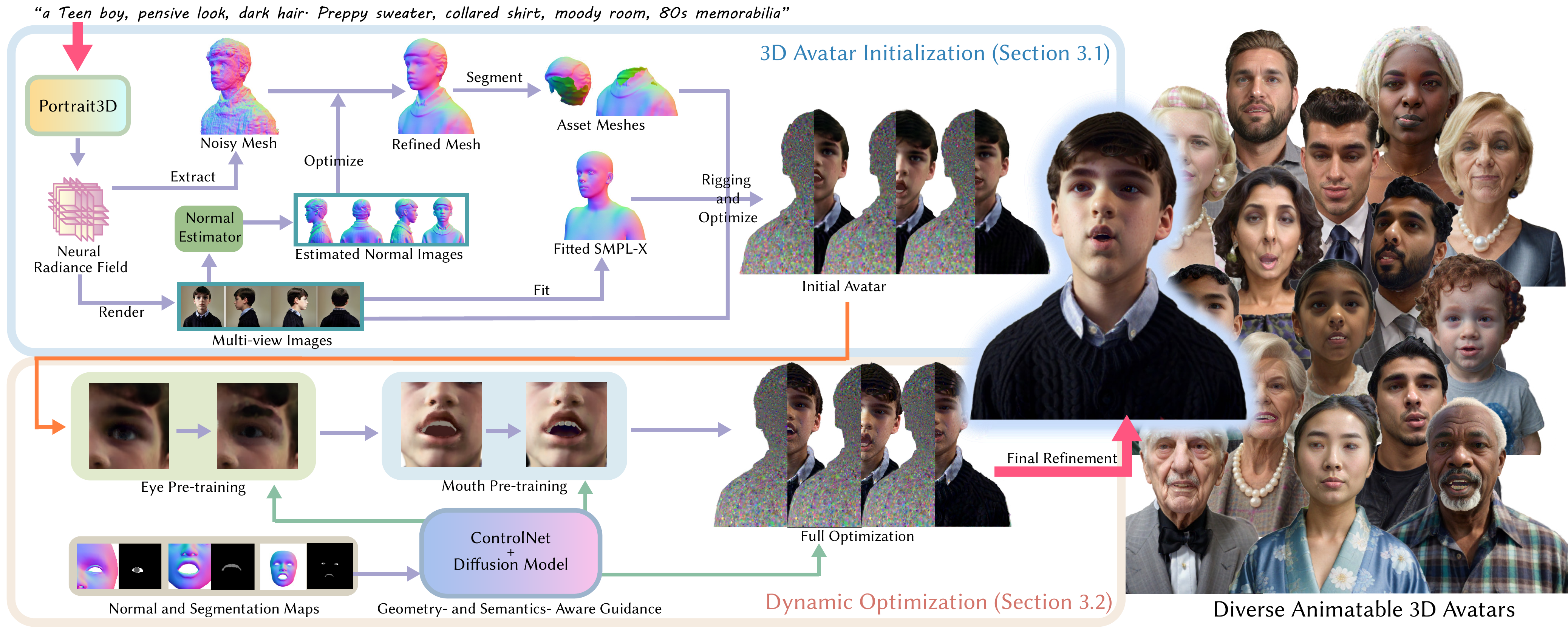}
  
  \caption{Overview of \modelname. Given an input text, the 3D Avatar Initialization stage (\Cref{sec: Initialization}) generates a well-defined initial avatar that provides appearance and geometry prior information, and is rigged to SMPL-X for animation. During the Dynamic Optimization stage (\Cref{sec: Optimization}), we optimize the avatar for dynamic poses and expressions using a 2D diffusion model and a ControlNet. 
  %we aim to generate highly detailed results while eliminating animation artifacts. 
  We first pre-train the eye and mouth regions, then optimize the full avatar and apply a refinement strategy to produce the final result. \modelname is able to generate avatars with diverse appearances, ethnicities, and ages.
  %\MP{I like it. Maybe we can add "Animatable 3D Avatar" to the Final avatar visualization} \MP{Maybe it makes sense to make clear what the output of Portrait3D is, and that you use it to render the Multi-View images and that the Normal Maps are estimated from the Multi-view-images. And maybe we can slightly change how 'Geometry- and semantics-aware guidance' is visualized.} \MP{much better now! Can you maybe also indicate that the SMPL-X model is fitted against the multi-view images? Also what is the eye icon on the bottom-right supposed to visualize? Can we maybe add sth like 'normal \& semantic maps' to the bottom left? I think we can simplify the NeRF triplanes to only one triplane icon}\onethousc{Have revised. } \MP{Perfect!}
  %\MP{can we maybe move this one page down so that this figure appears on the same page that we start the method section on? (not a strong opinion though)}
  }
  \label{fig: overview}
\end{figure*}

In this paper, we present a novel framework, \textit{\modelname}, for creating animatable 3D avatars from text descriptions, achieving realistic appearance and geometry while maintaining accurate alignment with the underlying parametric mesh. 
We choose 3DGS \cite{3DGS} for our 3D representation due to its efficient rendering capabilities and animation flexibility.
As shown in \Cref{fig: overview}, our method consists of two stages: an initialization stage that eliminates appearance ambiguities by creating a well-defined initial avatar, and a dynamic optimization stage that produces highly detailed results and eliminates animation artifacts. 
%
% \MP{isnt the intuition: we obtain a high-quality static avatar from the initialization stage but when we try to animate it with the SMPLX-rig, artifacts occur that we have to fix in stage 2? If so, I would clearly write it like that here} \onethousc{Sure. See above and below}
%
During the first stage, we initialize our avatar from the static text-to-3D model Portrait3D~\cite{portrait3d} which provides high-quality appearance and geometry information.
Specifically, we fit the SMPL-X model \cite{smplx} to the initialized avatar- to enable animation in the second stage of our optimization procedure later on. 
To handle hair and clothing, which are not modeled by SMPL-X, we extract a noisy mesh from the static avatar and refine it using estimated normal maps. The refined mesh is then segmented into hair and clothing components using pretrained semantic segmentation networks.
Then we texturize the SMPL-X, hair, and clothing meshes with 3D Gaussians and optimize their appearance features using multi-view images rendered from the static Portrait3D avatar. 
We adopt the rigging of SMPL-X to animate the avatar.
However, animating the avatar with SMPL-X rigging introduces artifacts due to its optimization in a static setting. To address this, we introduce the second stage
%that generates highly detailed results while eliminating animation artifacts.
in which we optimize the avatar for dynamic poses and expressions using a 2D diffusion model. 
% \MP{isnt the intuition: we obtain a high-quality static avatar from the initialization stage but when we try to animate it with the SMPLX-rig, artifacts occur that we have to fix in stage 2? If so, I would clearly write it like that here}
%
Specifically,
%to align with the SMPL-X model, 
we train a ControlNet \cite{controlnet} to provide geometry- and semantics-aware guidance based on the SMPL-X model's normal and segmentation maps as conditions. 
For challenging areas with complex geometry and frequent occlusions, such as the eyelids and mouth interiors, we introduce a pre-training strategy to refine color details. For the eye region, we leverage the diffusion model and ControlNet to generate a refined eye image, which is used to optimize the eye area of the 3D avatar. For the mouth region, we calculate the Interval Score Matching (ISM) loss~\cite{luciddreamer} using the diffusion model and ControlNet to refine the mouth area.
% \MP{if you mention this pretraining here, you have to explain what happens there as well}
% 
The full 3D avatar is then optimized with the ISM loss. {To eliminate minor artifacts and enhance the final quality, we further apply a refinement strategy by optimizing the 3D avatar with images refined by the diffusion model.} 
% \MP{here again: if you mention the refinement strategy here, you have to briefly summarize what is going on}
%
We demonstrate that our method can generate high-quality, animatable 3D avatars from text, achieving superior results in challenging areas such as the mouth interiors and eyes, and establishing robust rigging relationships with the corresponding parametric model.

In summary, our work makes the following key contributions:
\begin{itemize} 
    \item A novel framework, \modelname, for generating animatable, text-based 3D avatars with superior synthesis quality and animation fidelity.   
    \item A novel initialization strategy for animatable 3D avatar generation that integrates geometry and appearance initialization, while establishing robust rigging relationships with the corresponding parametric model.  
    \item A new geometry- and semantics-aware 3D avatar optimization method for dynamic poses and expressions, leveraging a ControlNet for robust guidance and improved alignment with the driving parametric model. 
    %A new geometry- and semantics-aware 3D avatar optimization method that uses a ControlNet for robust guidance and improved alignment with the driving parametric model. 

\end{itemize}

\section{Related Work}
\label{sec: Related Work}

\subsection{Animatable 3D Avatar Reconstruction}
3D avatar reconstruction typically utilizes captured real-world data to ensure accurate and realistic modeling. 
Instead of using parametric meshes to represent 3D avatars \cite{flame,smplx,bfm,Sphere_Face_Model,LNPH}, which could integrate well into industrial rendering pipelines, neural rendering techniques offer a promising approach for achieving more realistic results. Existing works have explored various methods, including neural textures \cite{neural-head-avatars}, NeRF \cite{RigNeRF,KeypointNeRF,DBLP:journals/tog/GaoZXHGZ22}, and 3D Gaussian Splatting (3DGS) \cite{GaussianAvatars,SplattingAvatar,GaussianAvatar} to enhance realism in 3D avatars. By conditioning the 3D representation with controllable parameters such as time, expression, and pose, dynamic details can be integrated into the avatar \cite{NerFACE, Gaussian-Head-Avatar, NPGA}, allowing for more dynamic and expressive models.
% \MP{also the works above are in 4D space!}\onethousc{Have changed to `dynamic details'} 
Prior models \cite{URAvatar,GGHead,HeadGAP} are also widely used for incorporating real-world data, enabling reconstruction and editing.

However, above methods all rely on real-world data as input, which can be cumbersome to obtain and raises potential ethical concerns. Our work builds upon the representation and rigging functions similar to GaussianAvatars, but instead of using real data, we employ a diffusion model to guide animatable 3D avatar generation from text prompts, eliminating the need for real data.

\subsection{Diffusion-based 3D Avatar Generation}
2D diffusion priors \cite{ddpm,latent-diffusion-model} hold great potential for 3D generation. While some methods still generate 2D images conditioned on 3D signals \cite{DiffusionRig,DiffPortrait3D}, DreamFusion \cite{DreamFusion} introduces Score Distillation Sampling (SDS), a novel approach for generating 3D content.
% To leverage 2D diffusion priors \cite{ddpm,latent-diffusion-model} for 3D content generation, DreamFusion \cite{DreamFusion} proposed a novel approach called Score Distillation Sampling (SDS). 
%
LucidDreamer \cite{luciddreamer} further improved this with Interval Score Matching (ISM), using deterministic diffusion trajectories and interval-based score matching for more robust updates during generation.
For text-based 3D avatar generation, the general pipeline involves applying SDS or its variants to 3D representations such as meshes \cite{seeavatar,teca,TADA,HumanNorm}, neural radiance fields \cite{AvatarVerse,portrait3d}, and 3DGS \cite{HeadStudio,HumanGaussian}. Most methods leverage geometric priors from parametric models, often overlooking appearance information. To mitigate issues of over-saturation and over-smoothness due to the lack of appearance priors, Portrait3D \cite{portrait3d} incorporates a GAN as a more robust joint prior. However, it still encounters noisy geometry and blurry artifacts.

To achieve animatable 3D avatar generation, parametric models such as SMPL-X \cite{smplx}, FLAME \cite{flame}, and imGHUM \cite{imGHUM} are widely used for their robust animation control, often with textures applied for appearance modeling \cite{TADA,seeavatar}.
Beyond 2D textures, 3D representations \cite{NeRF,Instant-ngp,3DGS} can also be rigged to parametric models \cite{HeadStudio,HumanGaussian,DreamHuman}, enabling detailed representation of complex features and efficient rendering. 
Rigging 3D representations to the surfaces of parametric models demands precise alignment with the underlying geometry.
However, in the generative setting, a 2D diffusion model alone lacks sufficient geometry- and semantics-aware guidance, often introducing ambiguities, resulting in low-quality results and animation artifacts.
Even HeadStudio \cite{HeadStudio}, which uses a landmarks-based ControlNet for additional guidance, exhibits artifacts because 2D landmarks do not provide enough geometric constraints.
Our method instead leverages dense normal- and semantic maps for conditioning which ensures better alignment and ultimately yields avatars with higher visual quality.
% Consequently, existing methods struggle with inadequate initialization and weak training guidance, leading to low-quality results and frequent artifacts in challenging areas like the eyelids and mouth.
% %

% \section{Preliminary}
% \onethous{
% In this paper, we adopt the rigging algorithm from GaussianAvatars \cite{GaussianAvatars} to facilitate animation.
% %
% GaussianAvatars uses the FLAME \cite{flame} model for animation, rigging Gaussian splats onto it for appearance rendering. For each triangle in the FLAME model, the mean position of its vertices, denoted as $\mathbf{T}$, is used as the origin of the local space. A matrix $\mathbf{R}$ is computed to describe the triangle's orientation in the global space, and a scalar $k$ represents the triangle's scaling factor.

% For each 3D Gaussian sampled on a triangle, its parameters are defined in the triangle's local space, including location $\mathbf{\mu}$, rotation $\mathbf{r}$, and anisotropic scaling $\mathbf{s}$. These parameters are converted to the global space during rendering:
% \begin{equation}
% \begin{split}
%     \label{eqn: GaussianAvatars}
%         \mathbf{r'} & = \mathbf{R}\mathbf{r} \\
%         \mathbf{\mu'} &= k\mathbf{R}\mathbf{\mu} + \mathbf{T} \\
%         \mathbf{s'} &= k\mathbf{s}.
%     \end{split}
% \end{equation}
% This approach enables normalized updates for each Gaussian in the local space and facilitates animation in coordination with the parametric model.
% }

\section{Methodology}
{In this section, we detail the two stages of our method: the 3D Avatar Initialization Stage (\Cref{sec: Initialization}), which eliminates appearance ambiguities by creating a well-defined initial avatar,
% \MP{I dont like the name 'Starting Avatar' to much. Could we maybe instead call it 'Initial avatar'?}\onethousc{Ok!},
and the Dynamic Optimization Stage (\Cref{sec: Optimization}), which resolves animation artifacts for dynamic poses and expressions.
%which generates highly detailed results and eliminates animation artifacts.
Please refer to \Cref{fig: overview} for an overview of our pipeline.
}

\begin{figure*}[t]
  \centering
  \includegraphics[width=0.99\linewidth]{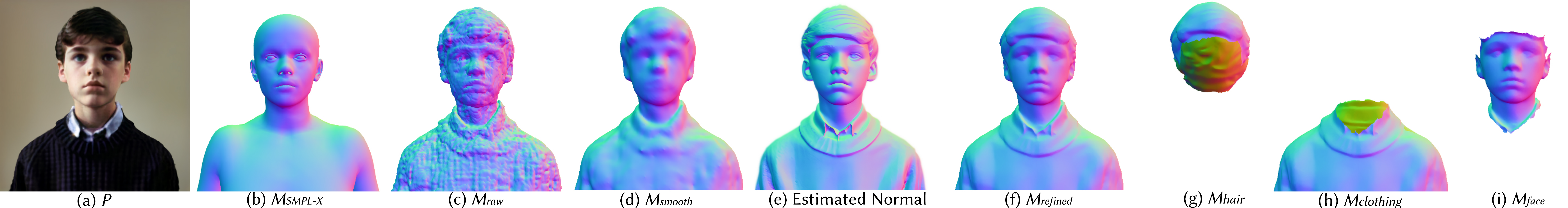}
  
  \caption{
     The visualization of (a) the static 3D avatar $P$ from Portrait3D,
     (b) the fitted SMPL-X model, (c) noisy mesh $M_{{raw}}$ extracted from $P$, (d) smoothed mesh $M_{{smooth}}$, (e) normal map estimated from the renderings of $P$, (f) $M_{{refined}}$ optimized against the estimated normal maps, (g) segmented hair mesh, (h) segmented clothing mesh, and (i) segmented face mesh.
   }
  \label{fig: mesh_vis}
\end{figure*}

\subsection{3D Avatar Initialization}
% \MP{Consider to add a preliminaries section for SMPL-X, Controlnet, Stable Diffusion, Portrait3D in which you briefly summarize them. Then Section 3 should be 'Method' with 3.1. Being something like 'Static Avatar Initialization' and 3.2. 'Dynamic Avatar Optimization'} \onethousc{I will include a sec for Portrait3D and reorganize Sections 3 and 4 into a single section. However, I’m still unsure whether a preliminaries section is necessary for SMPL-X, ControlNet, and Stable Diffusion, as these are relatively well-known concepts in the field. In my SIG 2024 submission, I did not include a preliminaries section, and the reviewers did not raise concerns about it. Additionally, we need to be mindful of the paper's length.} \onethousc{We can find a time to discuss this issue :)} \MP{makes sense}
\label{sec: Initialization}
The goal of the 3D Avatar Initialization Stage is to establish an initial 3DGS avatar with robust geometry and appearance, rigged to the SMPL-X model with semantic alignment.
We start with a static 3D avatar generated by Portrait3D~\cite{portrait3d} from the input text. 
First, we fit an SMPL-X model to the Portrait3D prediction and generate detailed meshes for hair and clothing, which cannot be represented with SMPL-X (\Cref{sec: Hair and Clothing Meshes Generation}). 
Then we sample 3D points from the obtained SMPL-X mesh, as well as the hair and clothing meshes, and establish rigging relations wrt. the SMPL-X model (\Cref{sec: Gaussian Splatting Geometry and Rigging Initialization}).
The sampled point cloud initializes our animatable 3DGS avatar representation, which is optimized using multi-view images obtained from the Portrait3D prediction (\Cref{sec: Gaussian Splatting Appearance Initialization}).

% We start with a static 3D avatar generated by Portrait3D \cite{portrait3d} and fit an SMPL-X model for animation.
% % 
% To generate high-quality hair and clothing meshes, which are not modeled by SMPL-X, we extract a noisy mesh from the Portrait3D predictions via MarchingCubes \cite{lorensen1998marching}, refine it with estimated normal maps, and segment it into hair and clothing components, as detailed in \Cref{sec: Hair and Clothing Meshes Generation}.  
% % 
% In \Cref{sec: Gaussian Splatting Geometry and Rigging Initialization}, we initialize the geometry and rigging of the 3DGS avatar. Specifically, asset Gaussians are initialized by sampling points from the hair and clothing meshes, while the remaining Gaussians are derived from points sampled on the fitted SMPL-X model. Rigging relationships are established semantically based on component membership.
% % 
% Following the initialization of geometry and rigging, in \Cref{sec: Gaussian Splatting Appearance Initialization}, we use the sampled point cloud to initialize the initial 3DGS avatar. Finally, the initial 3DGS avatar is optimized using multi-view images rendered from the static avatar.

\subsubsection{SMPL-X Optimization and Asset Mesh Generation}
\label{sec: Hair and Clothing Meshes Generation}
Given an input text prompt $y$, we first use Portrait3D to generate a static 3D avatar represented by a neural radiance field, denoted as $P$ (\Cref{fig: mesh_vis} (a)).  
Next, we apply multi-view head tracking \cite{VHAP} to obtain a fitted SMPL-X model ($M_{{SMPL-X}}$) from $P$, as shown in \Cref{fig: mesh_vis} (b).
While $M_{{SMPL-X}}$ can guide avatar animation, it cannot model assets like hair and clothing, which are crucial for the realism of the 3D avatar. 
To address this limitation, we propose to generate high-quality hair and clothing meshes from the 3D avatar $P$ to provide additional geometric information for these assets.
We extract $P$'s geometry as a raw mesh (denoted as $ M_{{raw}}$) using Marching Cubes \cite{lorensen1998marching}, as shown in \Cref{fig: mesh_vis} (c) and apply Laplacian smoothing to obtain $ M_{{smooth}} $ (\Cref{fig: mesh_vis} (d)). 
Then we render a set of multi-view images $ \{I_{{raw}}^i \mid i = 0, \cdots, N-1\} $ from $P$ and employ the normal estimator from Unique3D \cite{Unique3D} to extract normal maps (\Cref{fig: mesh_vis} (e)) from $ \{I_{{raw}}^i \} $.
We then optimize $ M_{{smooth}} $ using the estimated normal maps, resulting in a high-quality refined mesh, $ M_{{refined}} $, with enhanced detail and reduced noise (\Cref{fig: mesh_vis} (f)). 
Details of the mesh optimization can be found in \Cref{{sec: Meshes Generation}}.

% \begin{figure}[t]
%   \centering
%   \includegraphics[width=0.99\linewidth]{mesh_seg.pdf}
  
%   \caption{Face revoting. (a) For the faces whose indices fall within the hair segmentation region, we assign a positive hair score, while the others receive a negative hair score. (b) We apply a similar process to the face segmentation. \MP{I dont think we need this figure}}
%   \label{fig: mesh_seg}
% \end{figure}

\paragraph{Asset Mesh Segmentation}
\label{sec: Mesh Segment}
Inspired by MeshSegmenter \cite{MeshSegmenter}, we utilize Face Revoting to segment hair and clothing mesh components from $M_{{refined}}$. 
For each image in $\{I_{{raw}}^i\}$, we apply Sapiens \cite{Sapiens} to obtain 2D face segmentation maps for face mesh extraction. Due to Sapiens' limited generalization on hair and clothing, we employ a separate hair segmentation model \cite{pytorch_hair_segmentation} for accurate hair masks.
%
% \MP{shortened this section a bit} \onethousc{Nice!} 
We obtain the face and hair meshes by segmenting the faces of $M_{refined}$ through projection onto the face and hair segmentation maps and averaging over all views. 
For clothing, all remaining non-hair and non-face regions are designated as the clothing mesh.
The resulting segmented asset meshes are visualized in \Cref{fig: mesh_vis} (g-i).

% Specifically, using a differentiable renderer, we obtain a face index map of $ M_{{refined}} $ from each viewpoint in ${I_{{raw}}^i}$. 
% As shown in Fig. \ref{fig: mesh_seg}, for the faces whose indices fall within the hair segmentation region, we assign a positive hair score, while the others receive a negative hair score. 
% We apply a similar process to the face region.
% %
% Next, we classify all faces in $M_{{refined}}$ into three categories according to their assigned scores: the hair part $M_{{hair}}$, the face part, and the remaining clothing part $M_{{clothing}}$, as illustrated in Fig. \ref{fig: mesh_seg_and_fit} (b-d).

% \begin{figure}[t]
%   \centering
%   \includegraphics[width=0.8\linewidth]{mesh_seg_and_fit.pdf}
  
%   \caption{
%      Mesh segment results of (b) hair, (c) clothing, and (d) face parts. \MP{in case of space limits: we can merge this with fig 3}
%   }
%   \label{fig: mesh_seg_and_fit}
% \end{figure}

\subsubsection{Rigged Point Cloud Initialization}
\label{sec: Gaussian Splatting Geometry and Rigging Initialization}
Inspired by GaussianAvatars \cite{GaussianAvatars}, we sample points from the surfaces of $M_{SMPL-X}$, $M_{hair}$, and $M_{clothing}$ and rig them to $M_{SMPL-X}$'s faces. 
The resulting rigged point cloud initializes the animatable positions of the 3D Gaussians in our avatar representation. %initial avatar's Gaussians. 
For $M_{SMPL-X}$, we can directly adopt the rigging from the faces that the points were sampled from. 
For each point sampled from $M_{hair}$ and $M_{clothing}$, we find the closest face on the scalp and body partition of the SMPL-X model respectively and adopt their rigging parameters. 
%
% \Cref{fig: rigging} visualizes this procedure for the hair region. 
%

% Inspired by GaussianAvatars \cite{GaussianAvatars}, we use a similar rigging method to attach Gaussians onto $M_{{SMPL-X}}$ to achieve animation.  
% We sample points from the surface of the $M_{{SMPL-X}}$, then rig them to their corresponding faces.
%
% However, only directly sampling Gaussians onto the SMPL-X surface fails to accurately capture the avatar's geometry that deviates from the SMPL-X surface, such as hair and clothing.

% To capture more accurate asset geometry, we incorporate the hair and clothing meshes for hair and clothing geometry initialization.
% %
% For example, in the case of the hair mesh, we first sample points from the surfaces of $M_{{hair}}$, as illustrated in Fig. \ref{fig: rigging} (a). 
% %
% For sampled points on $M_{{hair}}$, as illustrated in Fig. \ref{fig: rigging} (b), we locate the nearest face on the scalp region of the SMPL-X model and rig the points onto the corresponding scalp face. 
% %
% Similarly, for clothing points on $M_{{clothing}}$, we follow the same process, rigging the points to the closest face on the body region of the SMPL-X model.
% %
% This process yields a point cloud rigged to the SMPL-X model, which is used to initialize the Gaussians' positions.

\subsubsection{Appearance Initialization}
\label{sec: Gaussian Splatting Appearance Initialization}
% \onethous{
% For the rigged point cloud, we query colors from the Portrait3D predictions, resulting in a colored point cloud with its rigging relationship to the fitted SMPL-X, as shown in \Cref{fig: gs_app} (a).}
% For the rigged point cloud, we query their colors from the Portrait3D predictions.
Next, we initialize our 3DGS avatar using the rigged point cloud and train it with the rendered images $\{I_{{raw}}^i\}$. 
Since mouth interior is completely invisible in the avatar initialization with neutral expression, we initialize the teeth with a generic proxy geometry and color.
For more details, please refer to \Cref{sec: Appearance Initialization}.
After this initialization process, our avatar exhibits high synthesis quality for a neutral expression, with all Gaussians semantically rigged to the SMPL-X model. We denote the trainable parameters of the initial 3DGS avatar as $\theta$.

% \MP{do you have a figure that shows the evolution from: rendering of portrait3d, rendering of initialized point cloud, rendering of Gaussian splats after static optimization? Would be nice to reference here} 
% %
% \onethousc{The rendering of Portrait3D and the rendering of Gaussian splats after static optimization appear quite the same, do you think we still need the figure? I added the figures of 1) rendering of initialized point cloud, 2) rendering of GS avatar after static optimization 3) the animated GS avatar with artifacts (for \Cref{sec: Optimization}). } 
% \MP{I guess you're right. The figure doesnt add as much insight as I had hoped for and I think we can drop it. Sorry for that :/ However, I think a figure similar to this can be very useful for section 4.2 with subfigures: a) Initial Avatar, b) Animated Starting Avatar with clear artifacts (maybe showing mouth opening), c) Animated Avatar after dynamic optimization of section 4.2) }\onethousc{ok!}

% \begin{figure}[t]
%   \centering
%   \includegraphics[width=0.99\linewidth]{gs_app.pdf}

%   \caption{
%      The face regions of (a) the initial avatar, (b) the animated initial avatar, and (c) the animated avatar after the optimization of \Cref{sec: Optimization}. We present zoomed-in views of the eye and mouth regions, showcasing the differences before and after optimization.
%      \MP{can we maybe have a sample where the differences pop out even more or have zoom-ins?}\onethousc{Have revised.}
%      }
%   \label{fig: gs_app}
% \end{figure}

\subsection{Dynamic Avatar Optimization}
\label{sec: Optimization}
While the avatar initialization of \Cref{sec: Initialization} yields an avatar that can be animated with the aligned SMPL-X model, in practice strong artifacts occur for novel expressions (see \Cref{fig: ablation} (a)). 
We identify two root causes for these artifacts: i)
for the eye region, minor misalignments between the eyelids and eyeballs of the underlying SMPL-X model with their corresponding Gaussians result in implausible rigging behavior, and ii) the initialization stage produces an avatar with a neutral expression and closed mouth, hence the interior of the mouth cavity is not represented well. 
We use a ControlNet that is conditioned on normal and semantic maps to fix these artifacts (\Cref{sec: ControlNet Training}).

Specifically, to address the underconstrained and poorly rigged eye and mouth regions in the initial avatar, we introduce a novel \textbf{pre-training strategy} (\Cref{sec: Eye and Mouth Region Initialization}), where high noise is applied to fully eliminate artifacts. Although the high noise may cause some blurriness, it establishes a robust initialization that benefits later stages.
Next, we apply \textbf{full optimization} (\Cref{sec: Full Optimization}), where the well-initialized but slightly blurry avatar from pre-training is refined using ISM with low noise. This step enhances details and corrects rigging inaccuracies while avoiding severe ambiguities.
While full optimization improves detail, it may introduce subtle high-frequency artifacts due to weak structural supervision in ISM gradients \cite{luciddreamer}. 
% These artifacts are removed in the \textbf{final refinement} stage (\Cref{sec: Final Refinement}) using SDEdit with low noise, which preserves fine details while producing clean, realistic results.
These artifacts are removed in the \textbf{final refinement} stage (\Cref{sec: Final Refinement}) using SDEdit with low noise, which preserves fine details while generating clean and realistic results.

\subsubsection{ControlNet Training}
\label{sec: ControlNet Training}
\begin{figure}[t]
  \centering
  \includegraphics[width=0.99\linewidth]{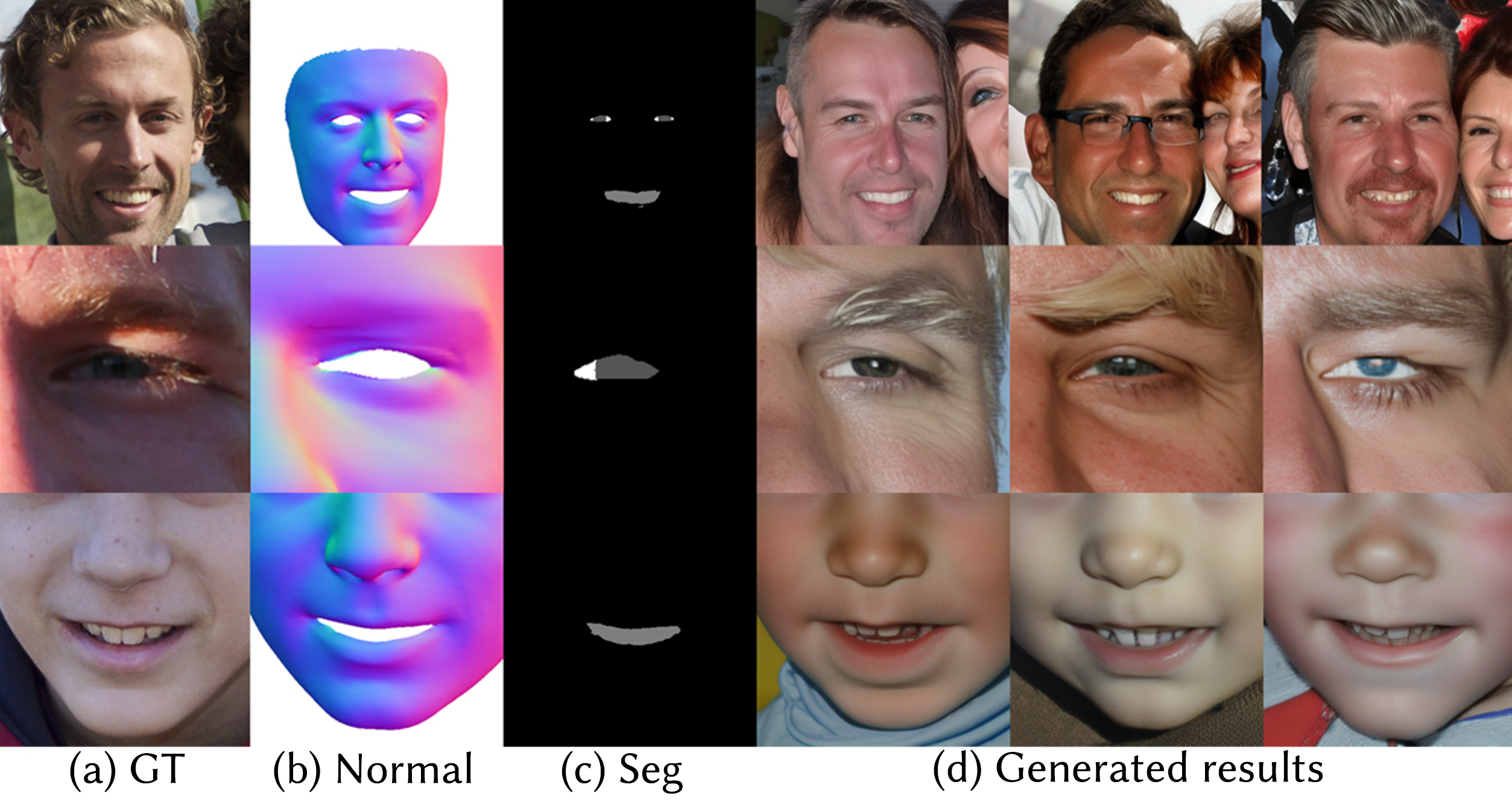}
  
  \caption{(a) The ground truth image from ControlNet's training dataset (originally derived from the FFHQ \cite{stylegan}). (b) Conditional normal maps. (c) Conditional segmentation maps. (d) Generated results using the corresponding inputs. }
  \label{fig: controlnet}
\end{figure}

We train a ControlNet to align the diffusion model’s guidance with the SMPL-X model.
Inspired by Joker \cite{joker}, we utilize normal maps as the conditional input of ControlNet. 
The training normal maps are extracted from RGB portrait images using a pretrained face reconstruction method \cite{DeepFace_recon}.
We supplement this control signal with segmentation maps for teeth, eyes and irises extracted with \cite{EasyPortrait,mediapipe} since these regions are underdetermined through the extracted normal maps (see \Cref{fig: controlnet}). 
To train the ControlNet, we construct a training dataset containing 453,385 high-quality paired RGB and conditional data, covering the face, mouth, and eye regions.
The details of the dataset and ControlNet training are provided in \Cref{sec: ControlNet Training}.
\Cref{fig: controlnet} illustrates the input conditionals and the corresponding results. 
%We denote the ControlNet trained on this dataset as $\epsilon_\mathcal{C}$.
During inference, we use the SMPL-X model’s rendered normal map and segmentation map as conditional signals for the ControlNet. 

% \begin{figure}[t]
%   \centering
%   \includegraphics[width=0.99\linewidth]{pipeline_eye_mouth.pdf}
  
%   \caption{The pipeline of eyes and mouth interiors pre-training. 
%   For the eye region, we generate aligned eye images based on the input conditional images, which are then used to optimize the avatar's eyes. For the mouth interiors, we compute gradients using ISM with the rendered images and their corresponding conditional inputs.
%   }
%   \label
%   {fig: pipeline_eye_mouth}
% \end{figure}

\subsubsection{Eye and Mouth Region Pre-training}
\label{sec: Eye and Mouth Region Initialization}
\begin{figure}[t]
  \centering
  \includegraphics[width=0.99\linewidth]{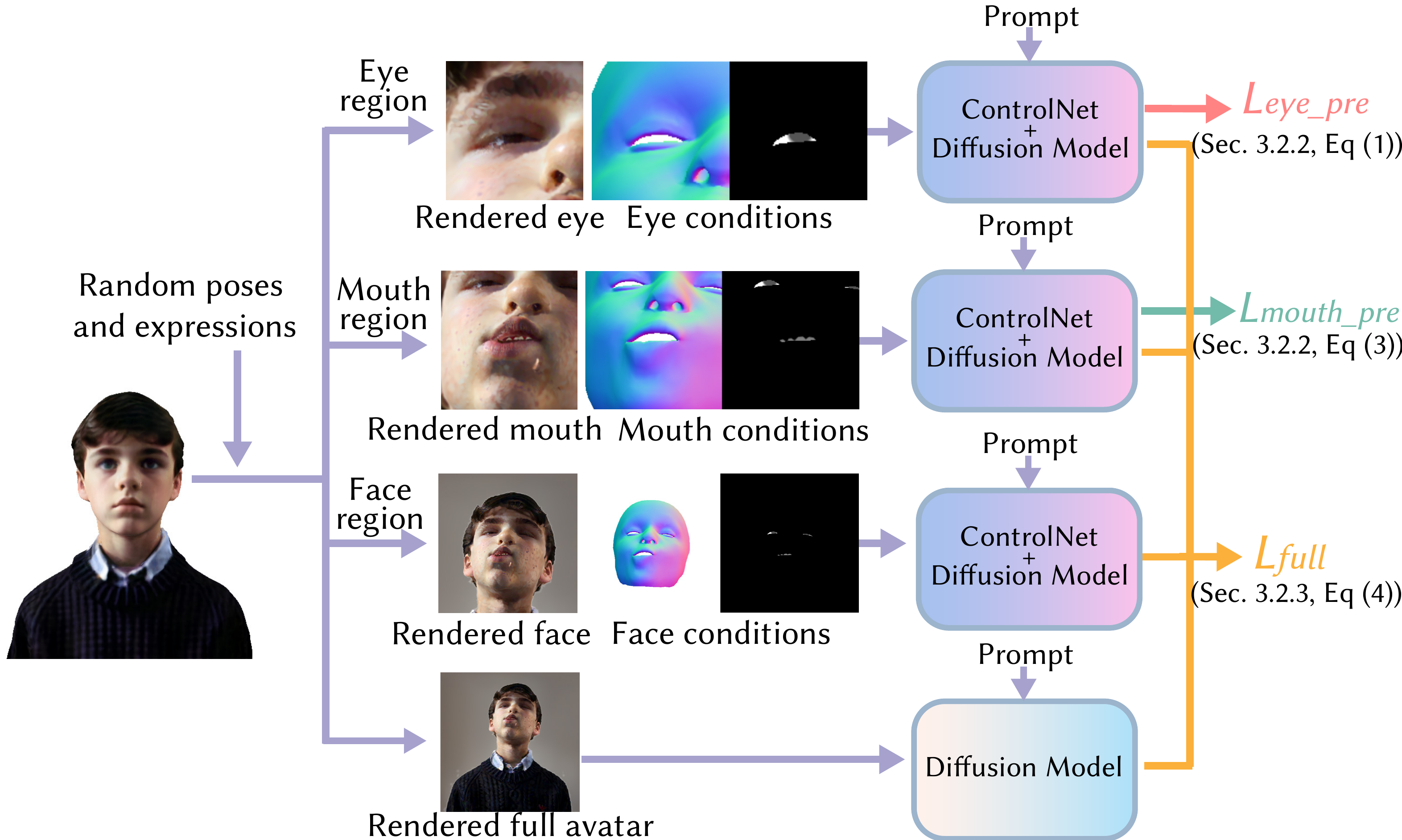}
  
  \caption{We optimize the eye region, mouth region, and full avatar sequentially, employing distinct loss functions at each stage. A Diffusion model together with a ControlNet conditioned on normal- and segmentation maps provide the guidance during optimization. Only for renderings of the full avatar, we omit the ControlNet and rely solely on the Diffusion model. %\MP{Watch out, typo: in the figure, the initialization losses are called $L_{eye\_init}$ and $L_{mouth\_init}$, but in the paper they are called $L_{eye\_pre}$ and $L_{mouth\_pre}$ respectively! Also fix the section and equation numbers.}\onethous{Done}
  }
  \label{fig: full_optimize}
\end{figure}

The eyes and the mouth interior are particularly challenging regions during animation. For the eyes, the exact alignment of the Gaussians for eyelids and eyeballs is critical, and the mouth interior is occluded in the neutral expression hence it is not well represented in the static avatar initialization. 
To solve this, we propose dedicated pre-training strategies for these regions. 
To refine the eye region, we leverage images generated by the ControlNet.
First, we randomly sample eyelid parameters, which control the opening and closing of the eyelids, and eye pose parameters, which define the gaze direction. These parameters are applied to the SMPL-X model to deform the 3D avatar accordingly.  
Next, the rendered eye region image and corresponding conditional inputs are processed using the ControlNet to perform SDEdit \cite{SDEdit}, generating a refined image to optimize the eye region of the avatar as follows:
\begin{equation}
\begin{split}
    \label{eqn: eye_init}
    L_{eye\_pre} =  L_2\left(I_{e}, 
    \text{SDEdit} (\epsilon_\mathcal{D}, \epsilon_\mathcal{C},I_{e},y,N_{e}, S_{e}) \right) ,
    \end{split}
\end{equation}
where $I_{e}$ denotes the rendered eye region image, and $N_{e}$ and $ S_{e}$ represent the eye region’s normal map and segmentation mask, respectively.  
$\epsilon_\mathcal{D}$ and $\epsilon_\mathcal{C}$ denote the pretrained Diffusion model and ControlNet model respectively. $y$ is the input prompt.
The refined eye image $ \text{SDEdit} (\epsilon_\mathcal{D}, \epsilon_\mathcal{C},I_{e},y,N_{e}, S_{e})$ is generated using the Diffusion model and ControlNet, with an editing strength set to 0.9. Please refer to the first row of \Cref{fig: full_optimize} for a visualization.

% Unlike the eye region, the mouth interior is completely invisible in the avatar initialization with neutral expression, hence we cannot obtain any color prior for this region. 
% %
% Therefore, rather than using generated images for refinement, we initialize the teeth with a generic proxy geometry and color 
Unlike the eye region, the mouth interior is only initialized with generic proxy geometry and color (\Cref{sec: Gaussian Splatting Appearance Initialization}). Therefore, instead of using generated images for refinement, we apply Interval Score Matching (ISM) \cite{luciddreamer} to optimize the mouth interior.
The ISM loss is defined as:
\begin{equation}
\begin{split}
\label{eqn: ism}
& \nabla_{\theta}\mathcal{L}_{ISM}(\theta, I, t,y,N,S)\\ 
       & \triangleq   \mathbb{E}_{t}\left[\omega(t)\left(\epsilon_\mathcal{D}\left(z_t , t, y, F_{ctrl} \right)-\epsilon_\mathcal{D}\left(z_t , s, \emptyset \right) \right) \frac{\partial z_0}{\partial I} \frac{\partial I}{\partial \theta}\right],\\
     & \text{where} \quad F_{ctrl} = \epsilon_\mathcal{C}(z_t, t,y, N, S).
\end{split} 
\end{equation}
Here the ISM loss $\nabla_{\theta}\mathcal{L}_{ISM}$ takes the trainable parameters $\theta$ of the 3DGS model, the rendered image $I$, the current time step $t$, the text y, the normal map $N$ and segmentation map $S$ as inputs, outputting the gradient to optimize the 3DGS model. $z_0$ is derived by feeding $I$ into the VAE encoder.
$s = t-\delta_T$ denotes the adjusted time step with a pre-defined inversion step size $\delta_T = 50$, and $\emptyset$ signifies the absence of conditionals.
$F_{ctrl}$ represents the features computed by the ControlNet $\epsilon_\mathcal{C}$, which are integrated into the Diffusion model $\epsilon_\mathcal{D}$ to provide guidance. 
%\MP{define what is $z_0$?} 

%
We sample open-mouth expressions from the NeRSemble dataset \cite{NeRSemble}, ensuring the visibility of the mouth interior and apply them to the SMPL-X model. 
The rendered mouth region, along with its associated conditional inputs, is processed through the diffusion model as follows: 
\begin{align}
\label{eqn: mouth_init}
& \nabla_{\theta}\mathcal{L}_{mouth\_pre}(\theta) = \nabla_{\theta}\mathcal{L}_{ISM}(\theta, I_m, t,y,N_m,S_m),
\end{align} 
where $I_{m}$ represents the rendered image of the mouth region, and $N_{m}$ and $S_{m}$ denote the mouth region’s normal map and segmentation map, respectively. This gradient formulation is used to optimize the mouth interiors of the 3D avatar, ensuring accurate refinement of this critical region. Please refer to the second row of \Cref{fig: full_optimize} for a visualization.

\subsubsection{Full Optimization}
\label{sec: Full Optimization} 
Following the pre-training of the eye and mouth region, we proceed to optimizing the full avatar. During this optimization, we randomly sample poses and expressions from the NeRSemble dataset \cite{NeRSemble}, along with random camera viewpoints. After applying these parameters to the SMPL-X model, the full optimization process is formally defined as follows:
% \begin{align}
% \label{eqn: full_optim}
% & \nabla_{\theta}\mathcal{L}_{full}(\theta) \nonumber\\ 
%        & \triangleq   \mathop{\sum}_{r \in \{e,m,f\}} \left(\mathbb{E}_{t}\left[\omega(t)\left(\epsilon_\mathcal{D}\left(z_t ; t,y, \epsilon_\mathcal{C}, N_{r}, S_{r} \right)-\epsilon_\mathcal{D}\left(z_t ; s, \emptyset \right) \right) \frac{\partial z_0}{\partial I_r} \frac{\partial I_r}{\partial \theta}\right] \right) \nonumber \\ 
%        &   +  \mathbb{E}_{t}\left[\omega(t)\left(\epsilon_\mathcal{D}\left(z_t ; t, y \right)-\epsilon_\mathcal{D}\left(z_t ; s, \emptyset\right) \right) \frac{\partial z_0}{\partial I_{full}} \frac{\partial I_{full} }{\partial \theta}\right].
% \end{align}
\begin{align}
\label{eqn: full_optim}
& \nabla_{\theta}\mathcal{L}_{full}(\theta)\\ 
       & \triangleq   \mathop{\sum}_{r \in \{e,m,f\}} \Big( \nabla_{\theta}\mathcal{L}_{ISM}(\theta, I_r, t,y,N_r,S_r) \Big) + \nabla_{\theta}\mathcal{L}_{ISM}(\theta, I_{full}, t,y).  \nonumber
\end{align}
In this formulation, we crop the eye region ($r=e$), mouth region ($r=m$), and face region ($r=f$) from the rendered full avatar $I_{full}$. Since these regions are significantly influenced by expression changes, we integrate the guidance provided by the ControlNet into their ISM losses. 
For optimizing the full rendering of the avatar however, i.e. the second term in \Cref{eqn: full_optim}, we discard the ControlNet and perform ISM only with the Diffusion model without conditioning it on normal- and semantic maps. 
We found that using the ControlNet for this scenario does not yield improvements since the image region that is influenced by expression changes (and therefore can be controlled with normal and semantic maps) is too small.  
% However, as the regions influenced by expression changes constitute only a small portion of the full avatar, adding ControlNet guidance results in only subtle improvements. 
Therefore, to speed up training, we omit ControlNet guidance for the full avatar renderings. Please refer to \Cref{fig: full_optimize} for a visualization.

\subsubsection{Final Refinement}
\label{sec: Final Refinement} 
 
Following the ISM-based optimization, we introduce a final refinement process to further enhance the quality of the results. Similar to the preceding optimization step, we render the avatar under random expressions, poses, and camera views, and subsequently refine the renders using SDEdit. These refined images are then employed to optimize the 3DGS model as follows:
\begin{equation}
\begin{split}
    \label{eqn: refine}
    L_{refine}  =  \mathop{\sum}_{r \in \{e,m,f,full\}}    & L_1\left(  \text{SDEdit}(\epsilon_\mathcal{D}, I_r,  y), I_r \right)  \\
    &+  L_{lpips}\left(  \text{SDEdit}(\epsilon_\mathcal{D},I_r, y), I_r \right)    ,
    \end{split}
\end{equation} 
where $\text{SDEdit}(\mathcal{D}, I_r, y)$ represents the refined image at different regions with an editing strength of 0.3. 
In this refinement process, the ControlNet is not used, as the relatively small editing strength (0.3) preserves the original structural integrity of the rendered image without requiring additional guidance. The terms $L_{1}$ and $L_{lpips}$ denote the L1 loss and the Learned Perceptual Image Patch Similarity (LPIPS) loss, respectively. These losses ensure that the refinement process balances pixel-level accuracy and perceptual quality.
Additionally, for the full avatar $r=full$ updating, we mask out updates on the non-face regions to prevent them from interfering with the more refined updates produced for the zoomed-in eye region ($r=e$), mouth region ($r=m$), and face region ($r=f$).
% \MP{Good!}

% \begin{table*}[t]

% \caption{Quantitative comparison results with SOTA methods. \textcolor{1st}{$\blacksquare$},\textcolor{2nd}{$\blacksquare$}, \textcolor{3rd}{$\blacksquare$} denote the 1st, 2nd, and 3rd places.  }
% \scalebox{0.99}{ 
% \begin{tabular}{@{}ccccccc@{}}
% \toprule
% \multirow{2}{*}{Method} & \multicolumn{2}{c}{Quality}           & \multicolumn{1}{c}{Sem Align} & \multicolumn{2}{c}{Geo Align}           & Human Preference \\ \cmidrule(l){2-7} 
%                         & HyperIQA $\uparrow$ & \multicolumn{1}{c}{DSL-FIQA $\uparrow$} & \multicolumn{1}{c}{CLIP $\uparrow$}  & Landmarks $\downarrow$ & AED $\downarrow$ & User Study $\uparrow$       \\ \midrule
% Ours          & \colorbox{1st}{61.2551} & \colorbox{1st}{0.6607}  & \colorbox{2nd}{0.2545} & \colorbox{1st}{0.0150} &   \colorbox{1st}{0.1154} \\
% HeadStudio    & {46.2231} & \colorbox{3rd}{0.3890} &\colorbox{1st}{0.2569} &\colorbox{3rd}{0.0250} &  {0.1675} \\
% TADA          & \colorbox{2nd}{57.0138} & {0.3400}  & {0.2321} &  {0.1436}& {0.1725} \\
% HumanGaussian & {25.2072} & {0.1072}  & \colorbox{3rd}{0.2531} & {0.0419}  & {0.2487} \\
% PortraitGen   & {39.9654} & {0.2964} & {0.2377} & {0.0266}  & \colorbox{3rd}{0.1575} \\
% GAGAvatar     &\colorbox{3rd}{49.8367} &\colorbox{2nd}{0.5645}  & {0.2518} & \colorbox{2nd}{0.0197}  & \colorbox{2nd}{0.1246} \\
% \bottomrule
% \end{tabular}
% \label{tab: results}
% }
% \end{table*}

\begin{figure*}[t]
  \centering
  \includegraphics[width=0.95\linewidth]{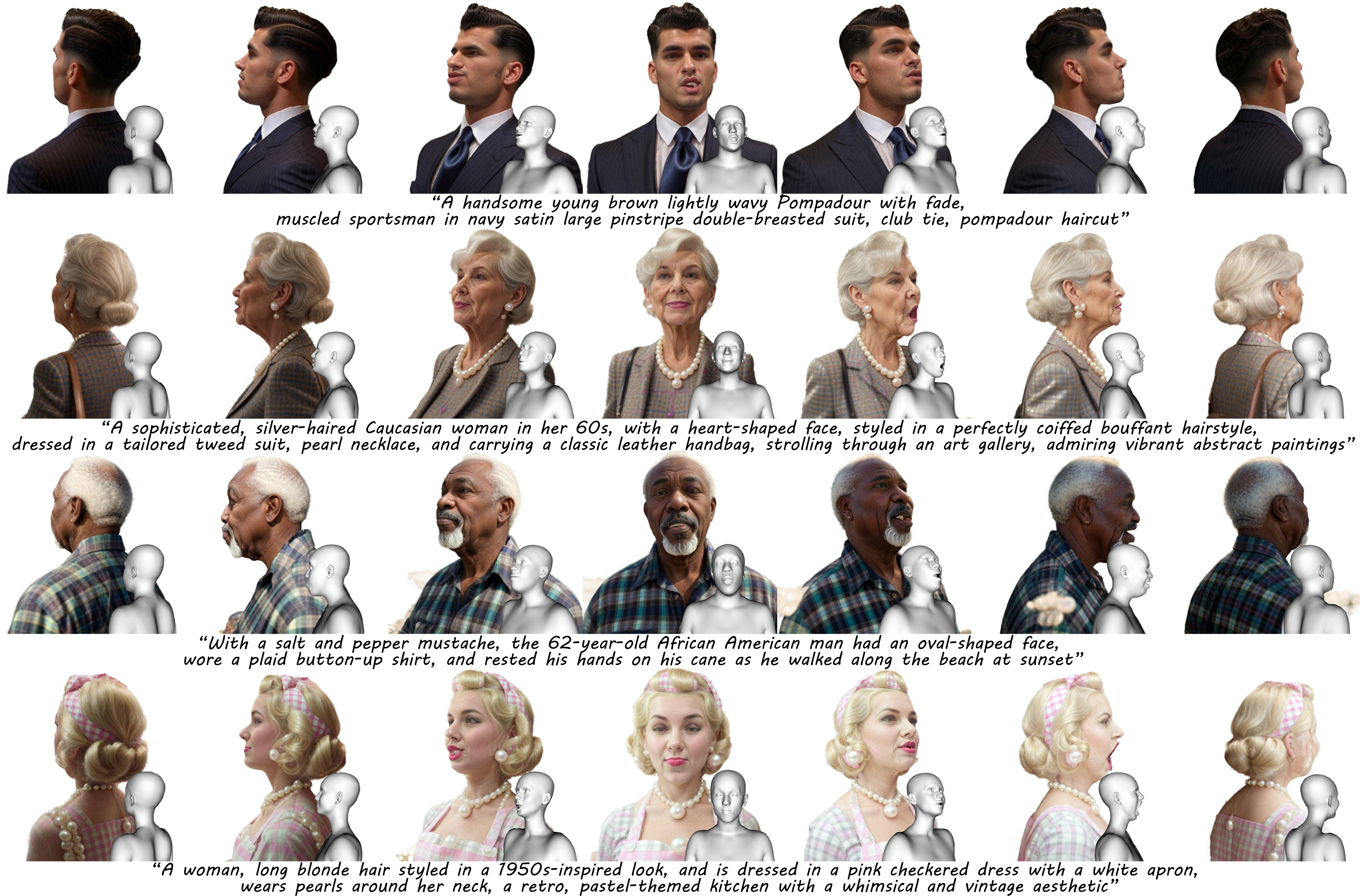}
  
  \caption{ 
    Generated results of our method. For each 3D avatar, we present rendered images with varying expressions and poses across different camera views, and the corresponding mesh for each avatar is shown at the lower right corner of each rendered image.
  }
  \label{fig: result-0}
\end{figure*}

\section{Results}
\label{sec: Results}

\subsection{Visual Results} 
\Cref{fig: result-0} showcases several generated 3D avatars rendered at various yaw angles, with random poses and expressions sampled from the NeRSemble dataset \cite{NeRSemble}.
To show the alignment accuracy, we also include corresponding SMPL-X model renderings alongside the avatars.
The results demonstrate the effectiveness of our method in generating highly detailed and realistic full-head 3D avatars, showcasing diverse appearances, ethnicities, and ages, along with realistic interior mouth and eyelid details that are difficult to achieve with existing methods. 
For dynamic visualizations, we utilize a head-tracking approach \cite{VHAP} to extract motion sequences from videos in the VFHQ dataset \cite{VFHQ}. The generated avatars are then driven by these motion sequences to show dynamic capabilities. Please refer to the demo video for dynamic results.

\subsection{Comparison}
We evaluate our method against SOTA approaches for animatable 3D avatar generation, including HeadStudio \cite{HeadStudio}, TADA \cite{TADA}, and HumanGaussian \cite{HumanGaussian}, using the same text prompt as input. Additionally, we compare our method with the SOTA 3D avatar editing approach, PortraitGen \cite{PortraitGen}, by employing the instruction, ``turn him/her into {\textit{text prompt}},'' to edit a 3D avatar using the same text prompt as that used for generation. 
We also include comparisons with 3DGS-based head reconstruction models, such as GPAvatar \cite{GPAvatar} and GAGAvatar \cite{GAGAvatar}, using the frontal image from Portrait3D as input.
%. Since our method builds upon results generated by Portrait3D, we utilize the frontal image from Portrait3D as the input for GAGAvatar to ensure a fair comparison.

\subsubsection{Qualitative Comparison}
We present the qualitative comparison results in \Cref{fig: comparison}.
To ensure a fair comparison, motion sequences extracted from the same reference video are used across all methods. Two frames are selected, and the rotating view for each frame is presented in the comparison image.
HumanGaussian, and PortraitGen face significant challenges in producing results that align with the driving frames, particularly in the eye and mouth regions. 
These limitations stem primarily from the absence of geometric supervision in their guidance frameworks. 
Even HeadStudio, which incorporates landmarks for expression alignment, struggles to fully eliminate misalignment. 
Moreover, the SDS-based methods (TADA, HumanGaussian, and HeadStudio) often generate unrealistic appearances. 
PortraitGen produces results with noticeable artifacts due to overfitting on the input video for reconstruction. 
GPAvatar and GAGAvatar often produce blurry outputs in hair and teeth regions, with the entire portrait appearing blurred at extreme camera angles. Furthermore,  their use of neural renderers to convert feature maps into RGB images frequently results in flickering artifacts.
In contrast, our method achieves superior results in challenging areas such as hair, eyes, and mouth. By integrating ControlNet, we significantly improve both alignment and controllability in complex regions and dynamic expressions. Additionally, our tailored initialization strategy enhances realism, preserving detailed and accurate appearances across all camera angles, including challenging back views.

% \begin{table}[t]

% \caption{Quantitative comparison results with SOTA methods. \textcolor{1st}{$\blacksquare$},\textcolor{2nd}{$\blacksquare$}, \textcolor{3rd}{$\blacksquare$} denote the 1st, 2nd, and 3rd places.}
% \scalebox{0.68}{ 
% \begin{tabular}{@{}ccccccc@{}}
% \toprule
% \multirow{2}{*}{Method} & Human Pref. & \multicolumn{2}{c}{Quality}           & \multicolumn{1}{c}{Sem Align.} & \multicolumn{2}{c}{Geo Align.}            \\ \cmidrule(l){2-7} 
%                        & User Study $\uparrow$  & HyperIQA $\uparrow$ & \multicolumn{1}{c}{DSL-FIQA $\uparrow$} & \multicolumn{1}{c}{CLIP $\uparrow$}  & Landmarks $\downarrow$ & AED $\downarrow$       \\ \midrule
% Ours       &   & \colorbox{2nd}{59.6879} & \colorbox{1st}{0.6426}  & \colorbox{1st}{0.2749} & \colorbox{1st}{0.0148} &   \colorbox{1st}{0.1265} \\
% HeadStudio &   & {46.7009} & {0.3819} &\colorbox{2nd}{0.2687} &\colorbox{3rd}{0.0263} &  {0.3136} \\
% TADA    &      & \colorbox{1st}{60.1467} & \colorbox{3rd}{0.4190}  & {0.2492} &  {0.1353}& {0.2162} \\
% HumanGaussian & & {23.8779} & {0.1123}  & \colorbox{3rd}{0.2661} & {0.0429}  & {0.2754} \\
% PortraitGen &  & {38.1552} & {0.2714} & {0.2485} & {0.0269}  & \colorbox{3rd}{0.1618} \\
% GAGAvatar   &  &\colorbox{3rd}{51.2861} &\colorbox{2nd}{0.5550}  & {0.2627} & \colorbox{2nd}{0.0199}  & \colorbox{2nd}{0.1267} \\
% \bottomrule
% \end{tabular}
% \label{tab: results}
% }
% \end{table}

\begin{table}[t]

\caption{Quantitative comparison results with SOTA methods. \textcolor{1st}{$\blacksquare$},\textcolor{2nd}{$\blacksquare$}, \textcolor{3rd}{$\blacksquare$} denote the 1st, 2nd, and 3rd places. ``Sem Align'' refers to semantic alignment, while ``Geo Align'' refers to geometric alignment.}
\scalebox{0.75}{ 
\begin{tabular}{@{}cccccc@{}}
\toprule
\multirow{2}{*}{Method} & \multicolumn{2}{c}{Geo Align} & \multicolumn{1}{c}{Sem Align} & \multicolumn{2}{c}{Quality}                      \\ \cmidrule(l){2-6} 
& Landmarks $\downarrow$ & AED $\downarrow$  & \multicolumn{1}{c}{CLIP $\uparrow$}   
                       & HyperIQA $\uparrow$ & \multicolumn{1}{c}{DSL-FIQA $\uparrow$}     \\ \midrule
Ours     & \colorbox{1st}{0.0148} &   \colorbox{1st}{0.1265}   & \colorbox{1st}{0.2749} & \colorbox{2nd}{59.6879} & \colorbox{1st}{0.6426}  \\
HeadStudio  &{0.0263} &  {0.3136}&\colorbox{2nd}{0.2687} & {46.7009} & {0.3819}  \\
TADA   &  {0.1353}& {0.2162} & {0.2492} & \colorbox{1st}{60.1467} & {0.4190}   \\
HumanGaussian   & {0.0429}  & {0.2754} & \colorbox{3rd}{0.2661}& {23.8779} & {0.1123}  \\
PortraitGen  & {0.0269}  & \colorbox{3rd}{0.1618}& {0.2485}  & {38.1552} & {0.2714}\\
GPAvatar &\colorbox{3rd}{0.0220}	&\colorbox{1st}{0.1265}	&0.2131 &49.4826&	\colorbox{3rd}{0.5157} \\
GAGAvatar& \colorbox{2nd}{0.0199}  & \colorbox{2nd}{0.1267}   & {0.2627} &\colorbox{3rd}{51.2861} &\colorbox{2nd}{0.5550}  \\
\bottomrule
\end{tabular}
\label{tab: results}
}
\end{table}

\subsubsection{Quantitative Comparison}  
\begin{figure*}[t]
  \centering
  \includegraphics[width=0.7\linewidth]{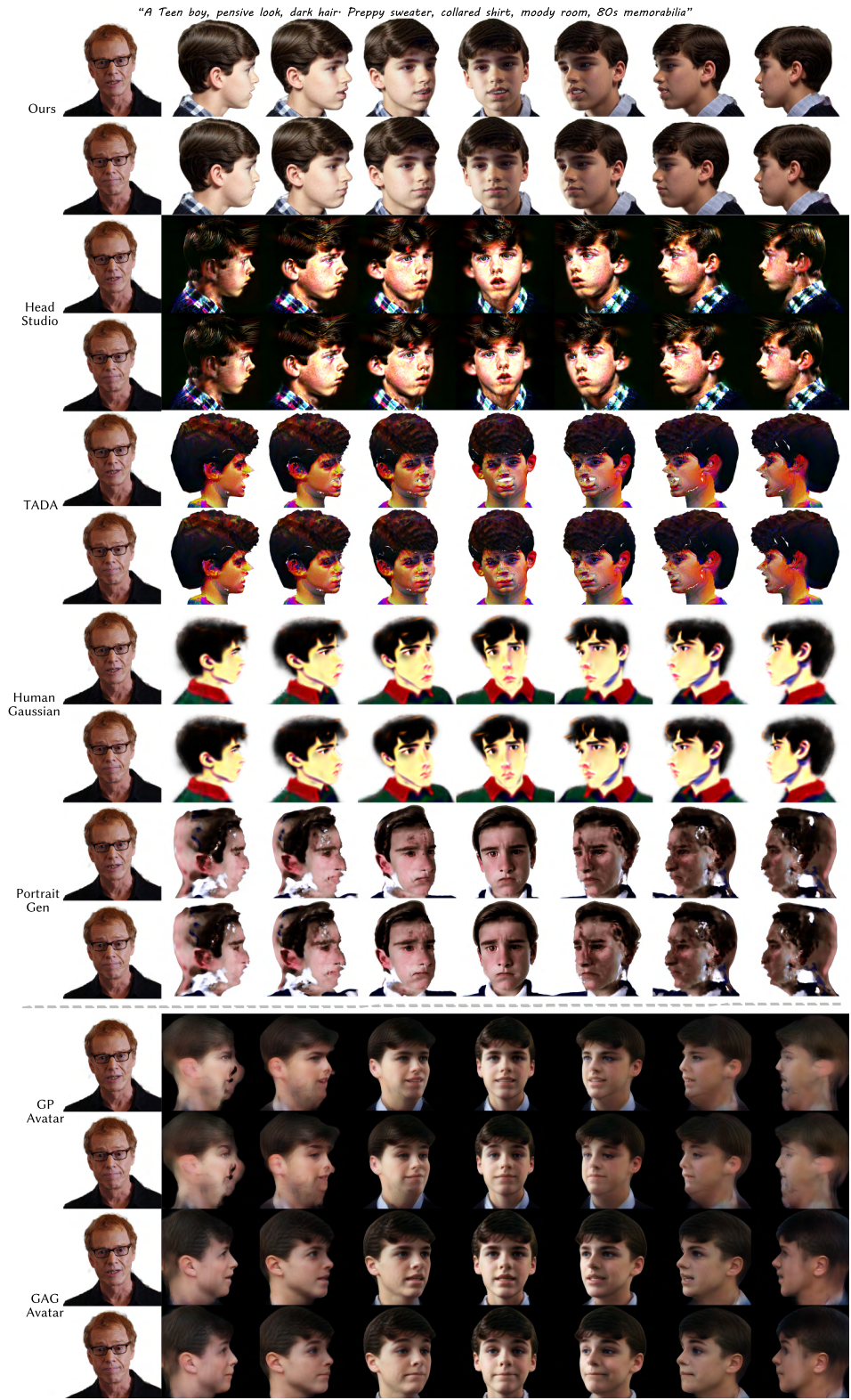}
  
  \caption{
     Comparison with HeadStudio \cite{HeadStudio}, TADA \cite{TADA}, HumanGaussian \cite{HumanGaussian}, PortraitGen \cite{PortraitGen}, GPAvatar \cite{GPAvatar}, and GAGAvatar \cite{GAGAvatar}. 
     While other methods take a text prompt as input (shown at the top), GPAvatar and GAGAvatar use an image as input. 
     The reference images are sourced from the video data in the VFHQ \cite{VFHQ} dataset.
     %\MP{I still think we can use one row per model and change expressions and camera poses simultaneously.} \onethousc{That may be a little confusing. In that case we would have two variables in a single row: camera view and expression. It is hard to say whether some artifacts are caused by the then misalignment with smpl-x or just due to the expression change in the sequence. } \MP{ok :)}
  }
  \label{fig: comparison}
\end{figure*}

For each method, we generate twenty avatars (ten male and ten female) using the same twenty prompts for all methods. To comprehensively evaluate the quality of the generated avatars, we render 100 random images for each avatar, with random camera views, and parameters randomly sampled from the NeRSemble dataset.
%
% To capture human preferences, we conduct a user study with XXX participants. Participants review videos rendered from 3D portraits generated by different methods and select the one with the highest overall quality and realism. The percentage of selections for each method is reported as the human preference score.
%
To evaluate the geometric alignment between the appearance of the generated avatars and their underlying meshes, we employ an off-the-shelf method \cite{STAR} to predict facial landmarks from the rendered images. We then calculate the deviations between these predicted landmarks and their corresponding points on the parametric models. For the Average Expression Distance (AED), we animate each avatar using a common reference video and apply a face capture method \cite{smirk} to estimate expression parameters from both the reference video and the generated avatars. The differences between these expression parameters are then computed to quantify expression alignment.
For semantic alignment, we measure the semantic consistency between the generated avatars and the input text by computing the CLIP similarity \cite{clip}.
To assess the quality of the generated avatars, we utilize HyperIQA \cite{HyperIQA}, a reference-free, general-purpose image quality assessment method, and DSL-FIQA \cite{DSL-FIQA}, a specialized facial image quality evaluation framework. 
%
%For image-level semantic alignment, we generate images from the corresponding text prompt using the diffusion model and compute average human visual similarity using DreamSim \cite{DreamSim}. 
% For text-level semantic alignment, we calculate the CLIP similarity \cite{clip} between the generated avatars and the input text prompts. 
% 
\Cref{tab: results} summarizes the quantitative evaluation results, illustrating that our method outperforms other approaches across most metrics. 
%
% Furthermore, in terms of general image quality, our method achieves performance comparable to TADA. \MP
%
We find that the general-purpose image quality score HyperIQA is insensitive to the strong artifacts of TADA and assigns a surprisingly high score which contradicts the qualitative observations in \Cref{fig: comparison}. We attribute this effect to limitations of HyperIQA's training data since the face-specific image quality metric DSL-FIQA aligns better with human preference.

\begin{figure*}[t]
  \centering
  \includegraphics[width=0.99\linewidth]{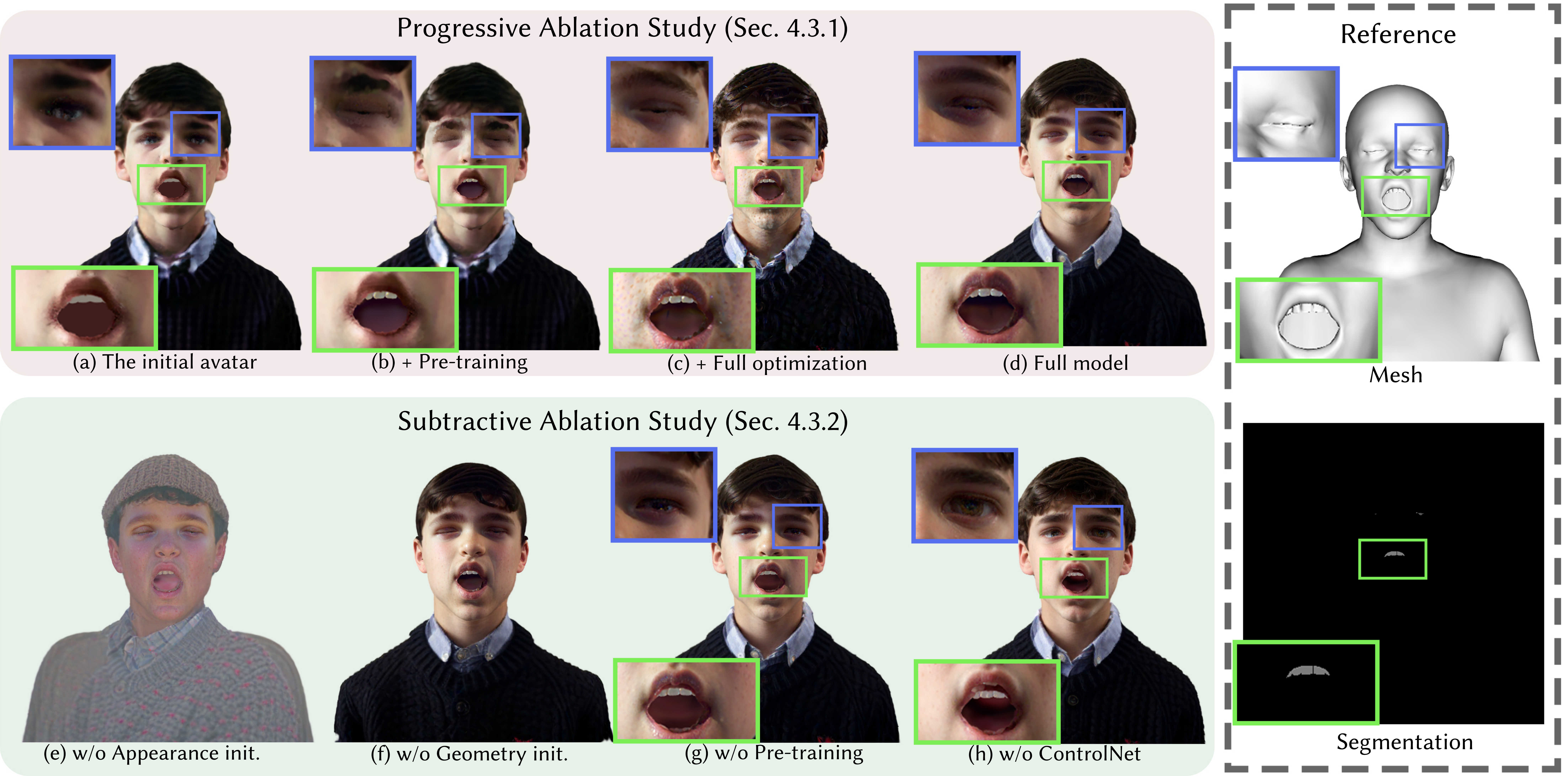}
  
  \caption{  
Ablation Study. We conduct two types of ablation studies: a progressive ablation study and a subtractive ablation study. The mesh renderings and the corresponding segmentation maps are shown on the right.
In the \textbf{progressive ablation}, we start with the initial avatar after the 3D Avatar Initialization Stage (a), then show the results after mouth- and eye pretraining (b), followed by the full optimization (c), after which refinement yields the final avatar of our full model (d). 
In the \textbf{subtractive ablation} study, we drop individual components of our pipeline while the rest is kept fixed.  
For results requiring additional focus on the eye and mouth regions, we include zoomed-in views for detailed examination.
    }
  \label{fig: ablation}
\end{figure*}

\subsection{Ablation Studies}
In this section, we discuss the effectiveness of our key components, presenting qualitative results in \Cref{fig: ablation}. 
We conduct a progressive ablation study (\Cref{sec: Progressive Ablation Study}) and a subtractive ablation study (\Cref{sec: Subtractive Ablation Study}).
In \Cref{sec: Quantitative Ablation Studies}, we further demonstrate the effectiveness of our dynamic optimization stage by conducting ablation studies that replace it with two alternatives. A detailed quantitative analysis is also provided in \Cref{sec: Quantitative Ablation Studies}.

\subsubsection{Progressive Ablation Study}
\label{sec: Progressive Ablation Study}
In the progressive ablation study, we examine the intermediate results of our pipeline, showcasing outputs from different stages to demonstrate how our approach incrementally improves the quality of the results. The progressive ablation study includes: 1) the results after the 3D Initialization stage, 2) the avatar after pre-training the mouth- and eye region, and 3) the avatar after full optimization but without refinement.

In \Cref{fig: ablation} (a-d) we present how the avatar quality is improved as we progress through the stages of our optimization pipeline. 
%With the integration of different stages of our method, the results in \Cref{fig: ablation} (e-h) become progressively more robust. 
After the 3D Initialization stage (\Cref{fig: ablation} (a)), the avatar exhibits animation artifacts such as inaccurate rigging, unrealistic color, and holes, as the avatar was optimized with a neutral expression only.
These issues are significantly mitigated by the pre-training strategy (\Cref{fig: ablation} (b)), though the appearance still lacks realism. 
After full optimization (\Cref{fig: ablation} (c)), additional details are added to the avatar, but unrealistic artifacts remain. 
These are resolved in the refinement process which yields the final avatar with highly realistic appearance and robust animation capabilities (\Cref{fig: ablation} (d)).
%After refinement is applied, the final result (\Cref{fig: ablation} (h)) appears highly realistic and aligns well with the mesh.

\subsubsection{Subtractive Ablation Study}
\label{sec: Subtractive Ablation Study}
In the subtractive ablation study, we individually remove 1) appearance initialization, 2) geometry initialization, 3) eye and mouth pre-training, and 4) ControlNet from our full model to evaluate their contributions.  

\paragraph{Appearance and Geometry Initialization}
As detailed in \Cref{sec: Initialization}, our method initializes both the appearance and geometry of the initial 3D avatar from predictions of Portrait3D. 
To underscore the importance of this initialization, we perform two ablation experiments: (1) training 3D avatars without appearance initialization, where color and opacity are instead set to default values, and (2) training 3D avatars by directly sampling and rigging Gaussians on the SMPL-X model, bypassing the initialization using asset meshes. 
The results shown in \Cref{fig: ablation} (e-f) reveal that without appearance initialization, the avatars exhibit unnatural color tones, with clothing appearing ambiguous and lacking detail. Similarly, without geometry initialization, the avatars fail to capture complex and realistic human features, resulting in less natural and lifelike outputs.

\paragraph{Eye Region and Mouth Region Pre-training} 
As discussed in \Cref{sec: Eye and Mouth Region Initialization}, achieving realistic results in the eye region and mouth interiors poses significant challenges, necessitating pre-training for these areas. To demonstrate the effectiveness of this approach, we optimize 3D avatars without pre-training for the eye region and mouth interiors.
As shown in \Cref{fig: ablation} (g), omitting pre-training affects the robustness and realism of these regions, as well as their alignment with the underlying geometry.

\paragraph{ControlNet} 
To evaluate the effectiveness of our ControlNet, we generate baseline avatars relying solely on the 2D diffusion model for guidance, excluding the use of ControlNet.
As illustrated in \Cref{fig: ablation} (h), the comparison reveals that, even with careful initialization and pre-training, the absence of ControlNet's robust guidance results in noticeable quality degradation: the eyes appear unnaturally large, the rigging is inaccurate, and the lips contain artifacts.

\section{Discussion}
\label{sec: Discussion}
 
While our method demonstrates significant advancements, it also has certain limitations that warrant further investigation.
First, the use of static Gaussian features constrains the adaptability of the Gaussians, with lighting and shadows baked into the fixed color values. This is especially problematic in areas like the teeth, which are sensitive to lighting changes, and prevents accurate rendering of dynamic details such as wrinkles.
Future work could address this by incorporating dynamic Gaussian attributes and facial data.
Second, the quality of mesh segmentation depends on 2D segmentation performance. Suboptimal results may lead to incorrect rigging or floating Gaussians. This could be mitigated by manual corrections or more robust segmentation models.
Finally, animation expressiveness is constrained by the blend shapes of the underlying 3DMM, limiting realistic rendering of long hair and complex garments, which often require physics-based simulation. 
Additionally, using video-extracted blend shapes causes lip sync artifacts, missing gaze animation, and less realistic expressions. Higher-quality results are expected with artist-designed or industry-grade blend shapes.

% Despite these limitations, our method lays the foundation for future improvements in 3D avatar generation and animation. By addressing these challenges, it is possible to further enhance the realism and versatility of the generated avatars.
%

\section{Conclusion}
\label{sec: Conclusion}

In this paper, we propose a novel framework, \modelname, for generating high-quality, animatable 3D avatars with realistic appearance and geometry from textual input. Our method comprises two key stages: an initialization stage to produce a well-defined initial avatar, and an optimization stage to refine this avatar for detailed and dynamic results. In the initialization stage, the initial avatar is created using a static model from the text-to-3D framework Portrait3D, augmented with carefully designed appearance and geometry initialization, and rigging computation. In the optimization stage, we further refine the initial avatar to resolve artifacts for dynamic poses and expressions using a 2D diffusion model and Interval Score Matching. To ensure accurate alignment with the SMPL-X model, we introduce a ControlNet that provides geometry- and semantics-aware guidance.
Extensive experiments demonstrated consistent improvements compared to previous methods. 
We hope our approach inspires further advancements in the field of 3D avatar generation, particularly in improving realism, adaptability, and expression fidelity.
 
% In addition to producing realistic and animatable avatars, our method excels in generating high-quality hair and clothing along with corresponding meshes. These assets can be utilized to enrich 3D hair and clothing datasets through reconstruction techniques, offering potential applications in fields such as virtual reality and gaming. We hope our approach inspires further advancements in the field of 3D avatar generation, particularly in improving realism, adaptability, and expression fidelity.

\begin{acks}
This work was supported by the SNSF project grant 200021 204840.
Malte Prinzler received funding from the Max Planck ETH Center for Learning Systems (CLS).
Xiaogang Jin was supported by the Key R\&D Program of Zhejiang (Grant No. 2024C01069) and the National Natural Science Foundation of China (Grant No. 62472373).
\end{acks}

\bibliographystyle{ACM-Reference-Format}
\bibliography{main-bibliography}

\clearpage
\appendix
\appendixpage

\setcounter{table}{0}   
\setcounter{figure}{0}
\setcounter{section}{0}
\setcounter{equation}{0} 
\renewcommand{\thetable}{A.\arabic{table}}
\renewcommand{\thefigure}{A.\arabic{figure}}
\renewcommand{\thesection}{A\arabic{section}}
\renewcommand{\theequation}{A\arabic{equation}}

In this supplement, we begin by discussing the implementation details of our method in \Cref{sec: Implementation Details}. 
{
In \Cref{sec: Quantitative Ablation Studies}, we show the additional ablation study on replacing dynamic avatar optimization and the qualitative ablation studies. 
}
In \Cref{sec: Gaussian Splatting Visualization}, we provide visualizations of the Gaussian Splats in our outputs. \Cref{sec: Comparison with Portrait3D} contains an additional qualitative comparison with Portrait3D. Then, we include further comparison results in \Cref{sec: Additional comparison} and showcase the additional visual results of our method in \Cref{sec: Additional Visual Results}.
{ 
Finally, we discuss the potential risks and corresponding countermeasures in \Cref{sec: Discussion}.

\section{Implementation Details}
\label{sec: Implementation Details}

\subsection{Meshes Generation}
\label{sec: Meshes Generation}
\subsubsection{Normal Map Estimation}
\label{sec: Normal Map Estimation}

To obtain the necessary geometry information, we render a set of multi-view images $ \{I_{{raw}}^i \mid i = 0, \cdots, N-1\} $ from the Portrait3D static avatar $P$. We then use an off-the-shelf normal estimator from Unique3D \cite{Unique3D} to extract normal maps from these multi-view images. 
To enhance details, we first use the pre-trained ControlNet-Tile \cite{controlnet} model to refine the quality of the multi-view images. ControlNet-Tile is a type of ControlNet model that works with a diffusion model, enabling it to refine images with enhanced details. 
For simplicity, we represent the combination of ControlNet-Tile and the diffusion model as $\mathcal{C}_{tile}$.
The detailed normal map generation process is as follows:
\begin{equation}
\begin{split} 
 I_{{normal}}^{i} &=    \mathcal{N} \left ( I_{{raw}}^{i}  \right)  =\mathcal{U} \left ( \mathcal{C}_{tile} \left ( I_{{raw}}^{i} \right) \right),
\end{split}
\end{equation}
where $\mathcal{U}$ denotes the normal diffusion model in Unique3D \cite{Unique3D}, and $\mathcal{N}$ denotes the normal estimator in our pipeline.
% \begin{equation}
% \begin{split} 
%  I_{{normal}}^{i} &=    \mathcal{N} \left ( I_{{raw}}^{i}  \right) ,
% \end{split}
% \end{equation}
% where $\mathcal{N}$ denotes the normal estimator. For more details, please refer to Sec. A.1 in the supplementary file.

\subsubsection{Mesh Geometry Optimization}
\label{sec: Mesh Geometry Optimization}
Since the Portrait3D avatar $P$ is represented by a neural radiance field, its geometry can be extracted as a raw mesh $M_{{raw}}$  using the marching cubes algorithm, as shown in \Cref{fig: mesh_optim_pipeline}.
To obtain a high-quality mesh, we use the estimated normal maps $I_{{normal}}^{i}$ to refine the raw mesh $ M_{{raw}}$. 

We first apply Laplacian smoothing to $ M_{{raw}} $, obtaining a smooth mesh $ M_{{smooth}} $, as shown in \Cref{fig: mesh_optim_pipeline}. $M_{{smooth}} $ is then refined by optimizing its vertex positions as follows:
\begin{equation}
\begin{split}
V^*  &= \mathop{\arg\min}_V \left( L_{normal} + L_{consistency} \right),   \\
L_{normal} &= 1 - \sum_{i} \cos \left( \mathcal{R}_{normal}(M_{{smooth}}, c^{i}), I_{{normal}}^{i} \right), \\
L_{consistency} &= \frac{1}{|\bar{\mathcal{F}}|} \sum_{j,k \in \bar{\mathcal{F}}} (1-\mathbf{n}_j \cdot\mathbf{n}_k ),\\
\end{split}
\end{equation}
where $L_{normal}$ and $L_{consistency}$ are the normal loss and the normal consistency regularization \cite{NDS}. $V$ is the vertex set of $M_{{smooth}} $, $ c^{i} $ is the camera associated with $ I_{{normal}}^{i} $, $ \mathcal{R}_{normal} $ is a differential mesh renderer that renders normal map from mesh, and $ \cos (\cdot, \cdot) $ denotes the cosine similarity. $\bar{\mathcal{F}}$  denotes the set of triangle pairs sharing a common edge, and $\mathbf{n}_j$ is the normal vector of triangle $j$. $L_{consistency}$ enforces normal consistency among neighboring faces for smoothness constraint across the surface.
The refined mesh $ M_{{refined}} $ is obtained from the optimized vertex set $ V^* $. 
%We show $ M_{{refined}} $ in \Cref{fig: mesh_vis} (e), which features more details and exhibits reduced noise.

\subsection{Teeth Mesh} 
To achieve high-fidelity rendering of detailed teeth segmentation maps, we integrate a teeth mesh\footnote{\hyperref[]{https://www.turbosquid.com/3d-models/realistic-human-jaws-and-tongue-3d-model-2014042}} into the SMPL-X model. This integration involves rigging the teeth mesh to the SMPL-X joint structure, attaching the upper teeth to the head joint and the lower teeth to the jaw joint. The tongue component is excluded, preserving only the detailed teeth mesh to enhance segmentation accuracy.

\begin{figure}[t]
  \centering
  \includegraphics[width=0.99\linewidth]{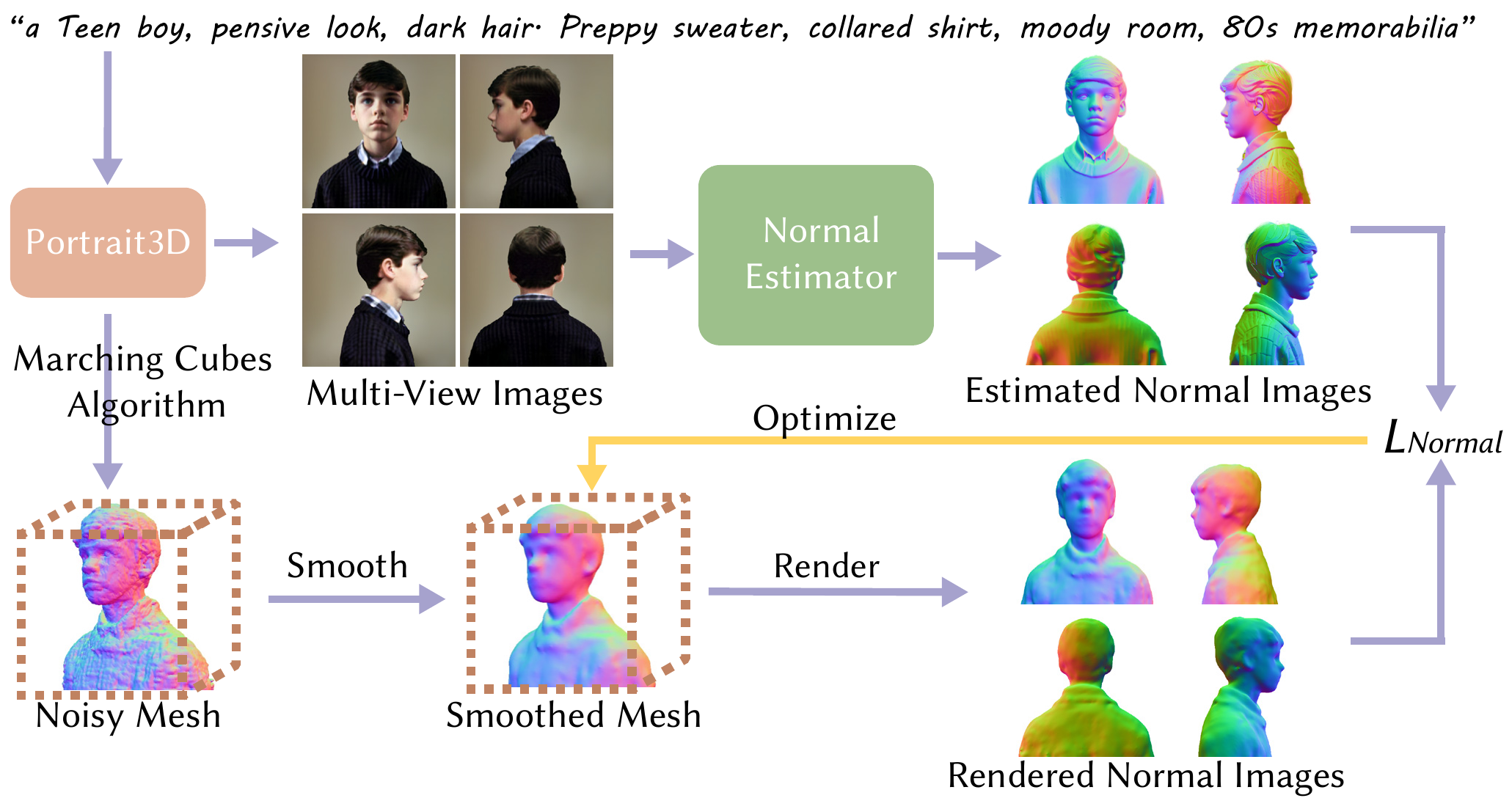}
  \caption{ 
     The pipeline of mesh optimization. Starting with the multi-view images generated by Portrait3D \cite{portrait3d}, we first employ a normal estimator to produce high-quality normal maps. Next, a noisy mesh is extracted from the Portrait3D output and smoothed using a Laplacian filter to reduce noise. Finally, we optimize the mesh by minimizing the discrepancy between the high-quality estimated normal maps and the rendered normal maps.
  }
  \label{fig: mesh_optim_pipeline}
\end{figure}

\subsection{Appearance Initialization} 
\label{sec: Appearance Initialization}
We use a sampled point cloud to initialize the 3DGS model. Instead of initializing the color of the sampled point cloud with default values, we leverage the color information from Portrait3D predictions to accelerate the appearance training process.
%
% Since Portrait3D utilizes a pyramid tri-grid structure, which is relative slow to query the feature of each point at each layer, we instead train a color field to query color for each point. 
%
Given the refined mesh $ M_{{refined}} $, we represent its texture as a hash-grid color field \cite{ngp}.
Next, we utilize multi-view images $ \{I_{{raw}}^i \}$ to optimize the color field.
We then feed the positions of the sampled point cloud into the optimized color field, resulting in a colored point cloud.

% Our appearance initialization process consists of two key steps: (1) training a color field to provide an initial color representation for the point cloud, and (2) initializing the 3D Gaussian Splatting (3DGS) using the colored point cloud and refining it through training with rendered images.

The color field, which is the texture of the refined mesh $ M_{{refined}} $, is represented as a hash-grid \cite{ngp}  $\Gamma_\delta(v) = \sigma$, where $ v \in V^* $ is the vertex position of $ M_{{refined}} $.
We utilize multi-view images $ \{I_{{raw}}^i \}$ to optimize the color field $ \Gamma_\delta $ as: 
\begin{equation}
\begin{split}
\delta^* &=  \mathop{\arg\min}_\delta L_2 \left(  \mathcal{R}_{{rgb}}(M_{{refined}}, \Gamma_\delta, c^{i}), I_{{raw}}^{i} \right) \ , 
\end{split}
\end{equation}
where $ \mathcal{R}_{{rgb}} $ is a differential mesh renderer that outputs an RGB image for the textured mesh.  We utilize a hash grid with a base resolution of 16, 12 levels, and a maximum resolution of 256. The color field is optimized with a learning rate of 0.01 for 600 iterations.

Next, we initialize a starting 3DGS avatar using this colored point cloud. For 3DGS training, we adopt the regularization terms from GaussianAvatars \cite{GaussianAvatars}, using a scale regularization weight of 1e4, a scale threshold of 0.2, position regularization of 1e5, and a position threshold of 1. Due to the color field initialization, we train the 3DGS for only 3,000 iterations—substantially fewer than required when training from scratch.

Since mouth interior is completely invisible in the avatar initialization with neutral expression, instead of training with multi-view images, we assign the 3D Gaussians on the teeth mesh a generic ivory white color ($R=141.6, G=133.8, B=122.4$), and the inner mouth skin a generic dark red color ($R = 64.0, G=30.5, B=29.5$), consistent with typical human inner mouth coloration.

\begin{figure}[t]
  \centering
  \includegraphics[width=0.99\linewidth]{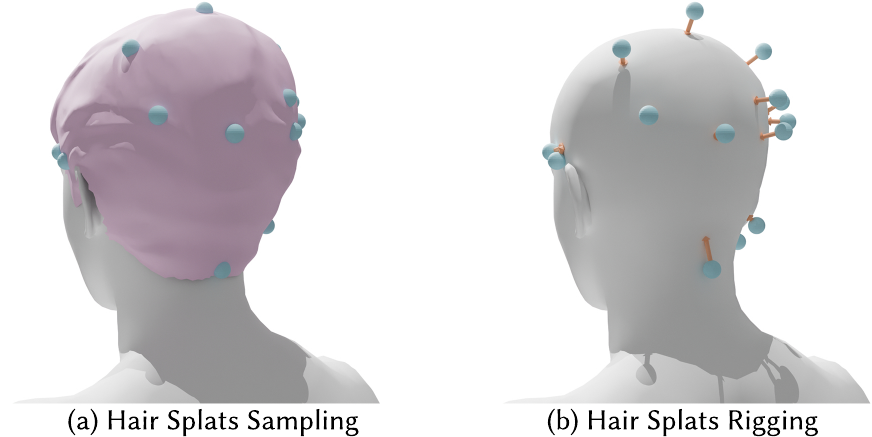}
  
  \caption{The hair Gaussians sampling and rigging. (a) Hair Gaussians (blue spheres) are sampled on the hair mesh (pink mesh). (b) The hair Gaussians are then rigged to the closest face on the scalp region of the SMPL-X model (gray mesh), with the rigging relationship indicated by the orange arrows. Note that we randomly sample portions of the hair Gaussians for clearer visualization. 
  \vspace{-10pt}
  }
  \label{fig: rigging}
\end{figure}

\subsection{Rigged Point Cloud Initialization}
We propose to generate high-quality hair and clothing meshes from the 3D avatar $P$ to provide additional geometric information for the assets. 
Specifically, in the case of the hair mesh, we first sample points from the surfaces of $M_{{hair}}$, as illustrated in \Cref{fig: rigging} (a). 
For sampled points on $M_{{hair}}$, as illustrated in \Cref{fig: rigging} (b), we locate the nearest face on the scalp region of the SMPL-X model and rig the points onto the corresponding scalp face. 
Similarly, for clothing points on $M_{{clothing}}$, we follow the same process, rigging the points to the closest face on the body region of the SMPL-X model.
This process yields a point cloud rigged to the SMPL-X model, which is used to initialize the Gaussians' positions.

\subsection{ControlNet Training} 
\label{sec: ControlNet Training}
To ensure accurate guidance for both the face, mouth, and eye regions, we construct a ControlNet training dataset that includes targeted data for each region.

For \textit{face} data, we utilize the FFHQ \cite{stylegan} and LPFF \cite{LPFF} (a large-pose variant of FFHQ) datasets. The text prompt for each image is extracted by BLIP \cite{blip}.
Using the 3D face reconstruction method \cite{DeepFace_recon}, we estimate normal maps as geometric conditional signals. 
We then apply Face Parsing \cite{EasyPortrait} to segment teeth and eye regions. Additionally, MediaPipe \cite{mediapipe} is used to track iris positions, providing further precision in gaze localization. 
% The visualization of these conditional signals is presented in \Cref{fig: controlnet} (b-c). 
 
\begin{figure}[t]
  \centering
  \includegraphics[width=0.99\linewidth]{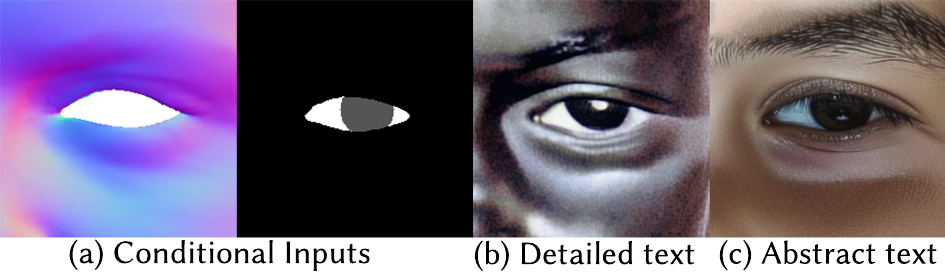}
  
  \caption{
      Using identical eye conditional inputs (a) and ControlNet, we observe that the detailed text prompt (b) — ``right eye region, a teen boy, pensive look, dark hair, preppy sweater, collared shirt, moody room, 80s memorabilia'' — produces lower-quality results compared to the more abstract text prompt (c) — ``right eye region, a boy''. 
  }
  \vspace{-10pt}
  \label{fig: detailed_text_vis_abstract}
\end{figure}

For \textit{eye} data, we first crop the eye regions from the face dataset. To augment the dataset with closed-eye variations, which are rare in in-the-wild portraits, we use LivePortrait \cite{LivePortrait}, a portrait animation method, to generate closed-eye variations from the FFHQ dataset. These closed-eye face images are then processed using a similar methodology to extract conditions, and the eye regions are cropped and added to the eye dataset.

To construct the \textit{mouth} dataset, we begin by cropping the mouth regions from the face dataset. To augment this dataset with a broader range of open-mouth variations, we incorporate additional images featuring open-mouth expressions sourced from the NeRSemble \cite{NeRSemble} dataset. These open-mouth face images are processed using a similar methodology to extract conditions, after which their mouth regions are cropped and integrated into the mouth dataset.

We construct the ControlNet training dataset using the face, eye, and mouth datasets, comprising 453,385 high-quality paired RGB and conditional data, covering these regions comprehensively.

The ControlNet is trained using the Realistic Vision V5.1 diffusion model\footnote{\hyperref[]{https://huggingface.co/SG161222/Realistic\_Vision\_V5.1\_noVAE}}. The training process takes approximately two days on an NVIDIA TITAN RTX GPU, with a batch size of 4 and a learning rate of
1e-4. During training, the probability of randomly dropping conditioning inputs is set to 0.1.
The conditional input consists of a concatenated normal map and segmentation map, resulting in a 4-channel input (3 channels for the normal map and 1 for the segmentation map). The resolution of training images is fixed at $512^2$. To ensure approximately balanced quantities of face, mouth, and eye data, we duplicate relevant samples. For data augmentation, we employ random resized cropping during training.

For ControlNet guidance on the \textit{face} region, we utilize the complete text prompt describing the full avatar (e.g., ``a teen boy, pensive look, dark hair, preppy sweater, collared shirt, moody room, 80s memorabilia''). However, for the \textit{mouth} and \textit{eye} regions, which typically lack person-specific features, we observe that detailed prompts degrade image quality, as demonstrated in \Cref{fig: detailed_text_vis_abstract} (b). Consequently, we use more abstract text prompts paired with region-specific prefixes for these areas (e.g., ``right eye region, a boy''), broadly categorizing the avatar, as shown in \Cref{fig: detailed_text_vis_abstract} (c).

\begin{table}[t]
\caption{ 
    The baseline models in our ablation studies. Used features are marked with  \Checkmark , and unused ones with  \XSolidBrush . N/A indicates that the model does not have a scenario to use the corresponding feature.
    The ablation studies are divided into four parts: our full model \textcolor{ours}{$\blacksquare$}, the progressive ablation study \textcolor{integrate}{$\blacksquare$}, the subtractive ablation study \textcolor{removal}{$\blacksquare$}, {and the ablation study on replacing the dynamic avatar optimization stage \textcolor{replace}{$\blacksquare$}.}
}
\scalebox{0.7}{ 
\begin{tabular}{@{}ccccccc@{}}
\toprule
Method & App. Init. & Geo. Init & Pre-training & Full Optim. & Refine & ControlNet \\ \midrule
 \rowcolor{ours} Ours  & \Checkmark &  \Checkmark &\Checkmark &\Checkmark &\Checkmark &\Checkmark \\ \cmidrule(l){1-7} 

 \rowcolor{integrate} Initial Avatar &\Checkmark &\Checkmark & \XSolidBrush&\XSolidBrush &\XSolidBrush & N/A \\  
\rowcolor{integrate} + Pre-training & \Checkmark&\Checkmark &\Checkmark & \XSolidBrush& \XSolidBrush& \Checkmark\\ 
\rowcolor{integrate} + Full Optimization &\Checkmark &\Checkmark &\Checkmark &\Checkmark & \XSolidBrush& \Checkmark\\ \cmidrule(l){1-7} 

\rowcolor{removal} - Appearance Init. & \XSolidBrush  &\Checkmark &\Checkmark & \Checkmark&\Checkmark & \Checkmark\\ 
\rowcolor{removal} - Geometry Init. &\Checkmark &\XSolidBrush &\Checkmark &\Checkmark &\Checkmark & \Checkmark\\
\rowcolor{removal} - Pre-training &\Checkmark &\Checkmark &\XSolidBrush &\Checkmark &\Checkmark &\Checkmark \\
\rowcolor{removal} - ControlNet  & \Checkmark &  \Checkmark &\Checkmark &\Checkmark &\Checkmark  & \XSolidBrush\\ \cmidrule(l){1-7} 

 \rowcolor{replace} {Refine} &{\Checkmark} &{\Checkmark} &\multicolumn{3}{c}{{Replaced with the final refinement}} & {\Checkmark} \\
\rowcolor{replace} {SR}  & {\Checkmark} &  {\Checkmark} &\multicolumn{3}{c}{{Replaced with super-resolution}}  & {N/A} \\

  \bottomrule
\end{tabular}
\label{tab: ablation-details}
}
\end{table}

\subsection{3D Avatar Optimization} 
In our pipeline, we employ the Realistic Vision V5.1 as the base diffusion model. The optimization process is conducted on an NVIDIA TITAN RTX GPU. For regularization, we adopt the terms introduced in GaussianAvatars \cite{GaussianAvatars}, with the following parameters: a scale regularization weight of 1e4, a scale threshold of 0.2, a position regularization weight of 1e-2, and a position threshold of 1.

During eye pre-training, the eye region is refined over 500 iterations.
During mouth pre-training, ISM is performed with 500 iterations, with the noise sampling level linearly decreasing from $t=750$ to $t=15$.
During full optimization, ISM is performed with 1,000 iterations, with the noise sampling level linearly decreasing from $t=300$ to $t=15$.
During final refinement, the full avatar is refined over 750 iterations.

\subsection{Runtime} 
The runtime for generating a 3D avatar is broken down as follows: the avatar initialization requires approximately 30 minutes, eye pre-training takes 20 minutes, mouth pre-training takes 25 minutes, ISM optimization requires 100 minutes, and the final refinement step takes 30 minutes. In total, the complete process to generate an avatar is approximately 3.5 hours. All experiments are conducted on an NVIDIA TITAN RTX GPU.

\begin{figure*}[t]
  \centering
  \includegraphics[width=0.99\linewidth]{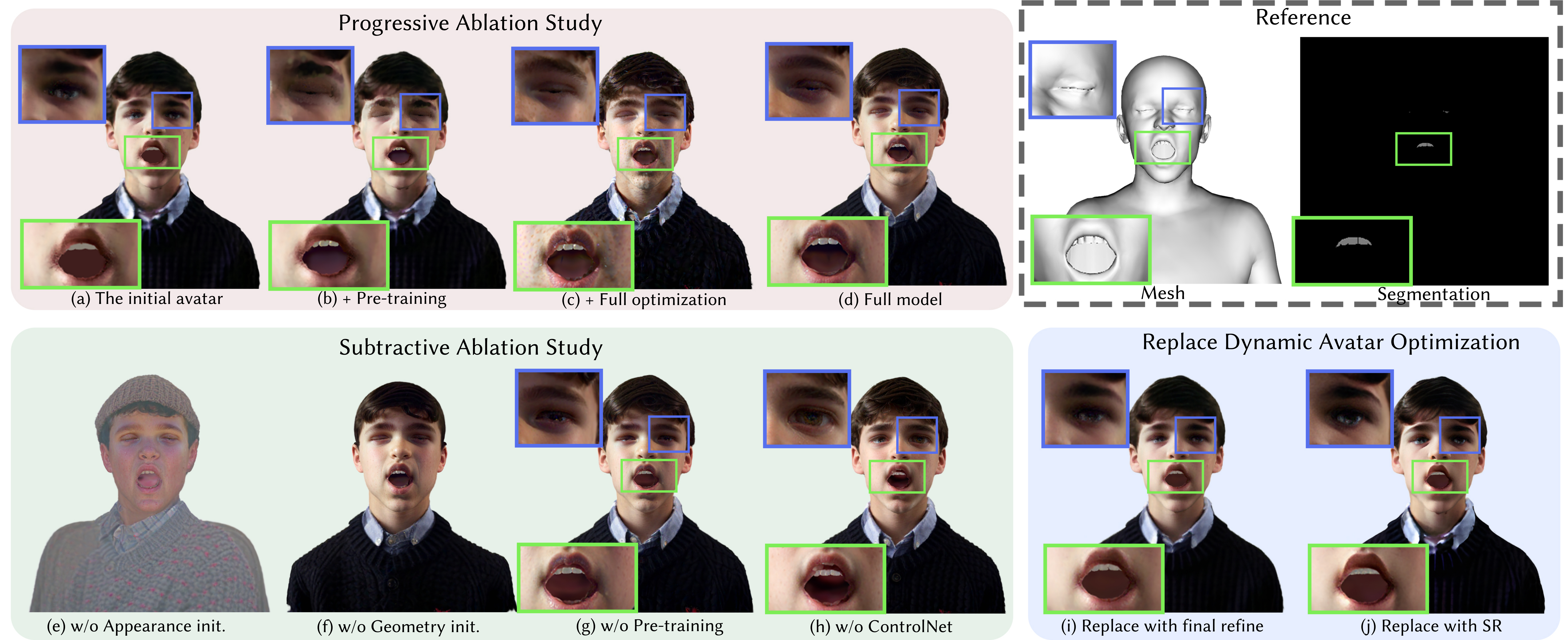}

  \caption{  
{Ablation Study. Here we conduct three types of ablation studies: a progressive ablation study, a subtractive ablation study, and an ablation study on replacing the dynamic avatar optimization stage. The mesh renderings and the corresponding segmentation maps are shown in the top right.
In the \textbf{progressive ablation}, we start with the initial avatar after the 3D Avatar Initialization Stage (a), then show the results after mouth- and eye pretraining (b), followed by the full optimization (c), after which refinement yields the final avatar of our full model (d). 
In the \textbf{subtractive ablation study}, we drop individual components of our pipeline while the rest is kept fixed. 
In the \textbf{replacement ablation study}, we replace the dynamic avatar optimization stage with either (i) final refinement or (j) super-resolution and present the resulting outputs.
For results requiring additional focus on the eye and mouth regions, we include zoomed-in views for detailed examination.}
    }
  \label{fig: ablation}
\end{figure*}

\section{Additional Ablation Studies}
\label{sec: Quantitative Ablation Studies}
\begin{table}[t]

\caption{Quantitative ablation studies. \textcolor{1st}{$\blacksquare$},\textcolor{2nd}{$\blacksquare$}, \textcolor{3rd}{$\blacksquare$} denote the 1st, 2nd, and 3rd places.  
 ``Sem Align'' refers to semantic alignment, while ``Geo Align'' refers to geometric alignment.
Here, we divide the results into four sections: our full model \textcolor{ours}{$\blacksquare$}, the progressive ablation study \textcolor{integrate}{$\blacksquare$}, the subtractive ablation study \textcolor{removal}{$\blacksquare$}, {and the ablation study on replacing the dynamic avatar optimization stage \textcolor{replace}{$\blacksquare$}.}
}
\scalebox{0.78}{ 
\begin{tabular}{@{}cccccc@{}}
\toprule
\multirow{2}{*}{Method} & \multicolumn{2}{c}{Geo Align} & \multicolumn{1}{c}{Sem Align} & \multicolumn{2}{c}{Quality}                      \\ \cmidrule(l){2-6} 
& Landmarks $\downarrow$ & AED $\downarrow$  & \multicolumn{1}{c}{CLIP $\uparrow$}   
                       & HyperIQA $\uparrow$ & \multicolumn{1}{c}{DSL-FIQA $\uparrow$}     \\ \midrule
\rowcolor{ours} Ours   & \colorbox{1st}{0.0148} &   \colorbox{1st}{0.1265} & \colorbox{2nd}{0.2749}  & \colorbox{3rd}{59.6879} &  \colorbox{3rd}{0.6426}  \\ \cmidrule(l){1-6} 

\rowcolor{integrate} Initial Avatar &{0.0167} &{0.1372} &{0.2727}  &{40.5922} & {0.3285} \\ 
\rowcolor{integrate} + Pre-training  &{0.0172} &{0.1355}  &{0.2688}  &{46.4792} &{0.3442} \\
\rowcolor{integrate} + Full Optimization &{0.0160} &\colorbox{3rd}{0.1270}  &{0.2687}  &{55.7962}  &{0.6008}\\ \cmidrule(l){1-6} 

\rowcolor{removal} - Appearance Init.  &\colorbox{2nd}{0.0150} &{0.1308} &{0.2530}  &{58.5367} &{0.6374} \\ 
\rowcolor{removal} - Geometry Init.&{0.0156} &\colorbox{2nd}{0.1266} &{0.2670}  &\colorbox{1st}{62.9362} &\colorbox{2nd}{0.6486} \\ 
\rowcolor{removal} - Pre-training  &\colorbox{3rd}{0.0154} &{0.1287} &\colorbox{3rd}{0.2747} &{58.5085} &{0.6302}\\  
\rowcolor{removal} - ControlNet  &{0.0181} &{0.1359} &\colorbox{1st}{0.2775} &\colorbox{2nd}{61.5838} &\colorbox{1st}{0.6587}\\ \cmidrule(l){1-6} 
 
\rowcolor{replace} {Refine}  &{0.0166}&	{0.1339}&	{0.2733}	&{45.6294}	&{0.4228}\\  
\rowcolor{replace} {SR}  &{0.0157}	&{0.1371}	&{0.2725}	&{51.4195}	&{0.5228}\\

\bottomrule
\end{tabular}
\label{tab: ablation}
}
\end{table}

{
\subsection{Ablation Study on Replacing Dynamic Avatar Optimization}
\label{sec: ablation Dynamic Avatar Optimization}
In this ablation study, we replace the entire dynamic avatar optimization stage with two alternative approaches, applying each directly to the initial avatar generated during the 3D avatar initialization stage. We show the details in \Cref{tab: ablation-details}.

\paragraph{Final Refinement}
As described in Section 3.2.4 of the main paper, we introduced a final refinement process to enhance result quality. In this experiment, we apply the refinement directly to the initial avatar. As shown in \Cref{fig: ablation} (i), using refinement alone leads to blurriness and inaccurate rigging.

\paragraph{Super Resolution}
We optimize the initial avatar using refined images produced by a super-resolution method, Real-ESRGAN \cite{Real-ESRGAN}. As shown in \Cref{fig: ablation} (j), while super-resolution slightly improves visual quality, the avatar still lacks detail and suffers from poor rigging.

}

\subsection{Quantitative Ablation Studies}
{
As mentioned in the main paper, we conduct two types of ablation studies: the progressive ablation study and the subtractive ablation study.
}
In the \textbf{progressive ablation study}, we examine the intermediate results of our pipeline, showcasing outputs from different stages to demonstrate how our approach incrementally improves the quality of the results. The progressive ablation study includes: 1) the results after the 3D Initialization stage, 2) the avatar after pre-training the mouth- and eye region, and 3) the avatar after full optimization but without refinement.
In the \textbf{subtractive ablation study}, we individually remove 1) appearance initialization, 2) geometry initialization, 3) eye and mouth pre-training, and 4) ControlNet from our full model to evaluate their contributions.  
{
In \Cref{sec: ablation Dynamic Avatar Optimization}, we additionally conduct \textbf{an ablation study
on replacing the dynamic avatar optimization stage}. We replace the entire dynamic avatar optimization stage with 1) the final refinement and 2) a super-resolution method, applying each directly to the initial avatar generated during the 3D avatar initialization stage.
}%
We show the details of these baselines in \Cref{tab: ablation-details}.

As shown in \Cref{tab: ablation}, we generate twenty avatars for each baseline, conducting quantitative experiments similar to the comparison section in our main paper to evaluate the individual contributions of the components in our framework.
Our full framework achieves the best geometric alignment, whereas removing ControlNet significantly weakens expression alignment. The intermediate results, including the initial avatar and the avatar after pre-training, also exhibit poor geometry alignment, highlighting the importance of our full optimization strategy.
Our full framework also delivers comparable performance on semantic alignment, while omitting appearance or geometry initialization reduces semantic alignment.
In terms of image quality, our model does not achieve the best performance, but the difference is minimal. The intermediate results, including the initial avatar and the avatar after pre-training, face significant quality degradation due to the absence of full optimization.
Note that the image quality metrics we use primarily evaluate image sharpness, they provide some indication of quality but cannot fully capture the realism of the face.

\begin{figure*}[t]
  \centering
  \includegraphics[width=0.7\linewidth]{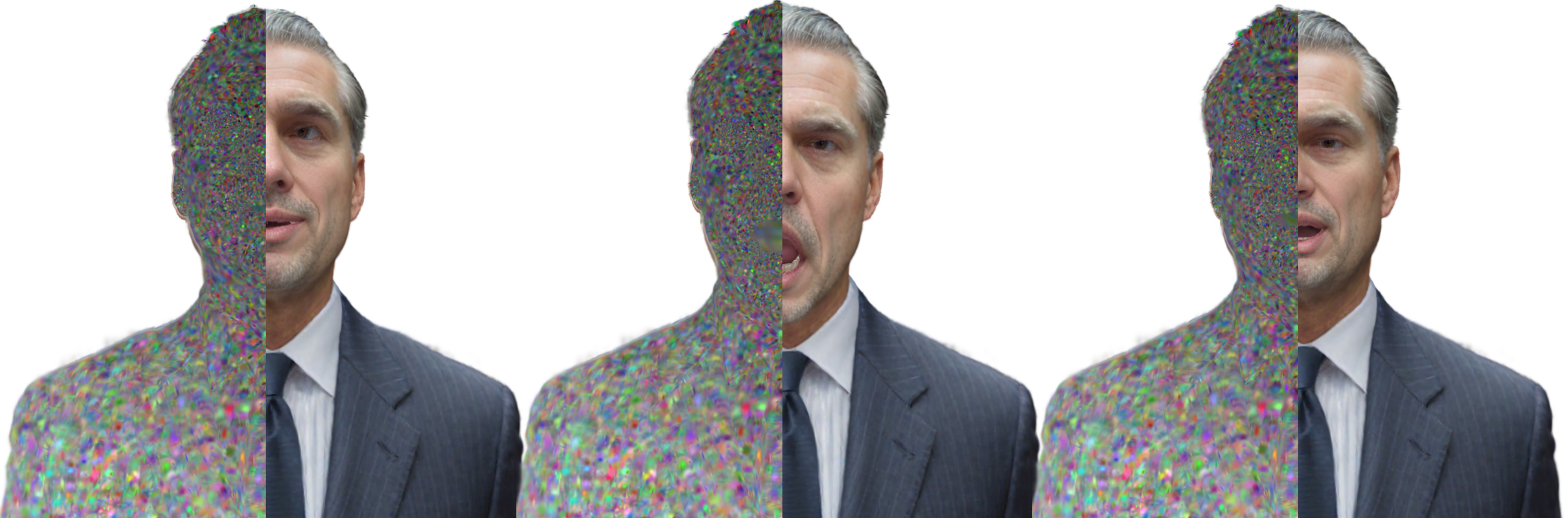}
  \caption{ 
      The figure showcases the original rendering of our results (right part of each rendering) alongside variations with randomly assigned colors (left part of each rendering).
  }
  \label{fig: gs-vis}
\end{figure*}

\section{Gaussian Splatting Visualization}
\label{sec: Gaussian Splatting Visualization}
To visualize the Gaussian primitives, \Cref{fig: gs-vis} presents renderings of our results, alongside variations where colors are randomly assigned.

\begin{figure*}[t]
  \centering
  \includegraphics[width=0.8\linewidth]{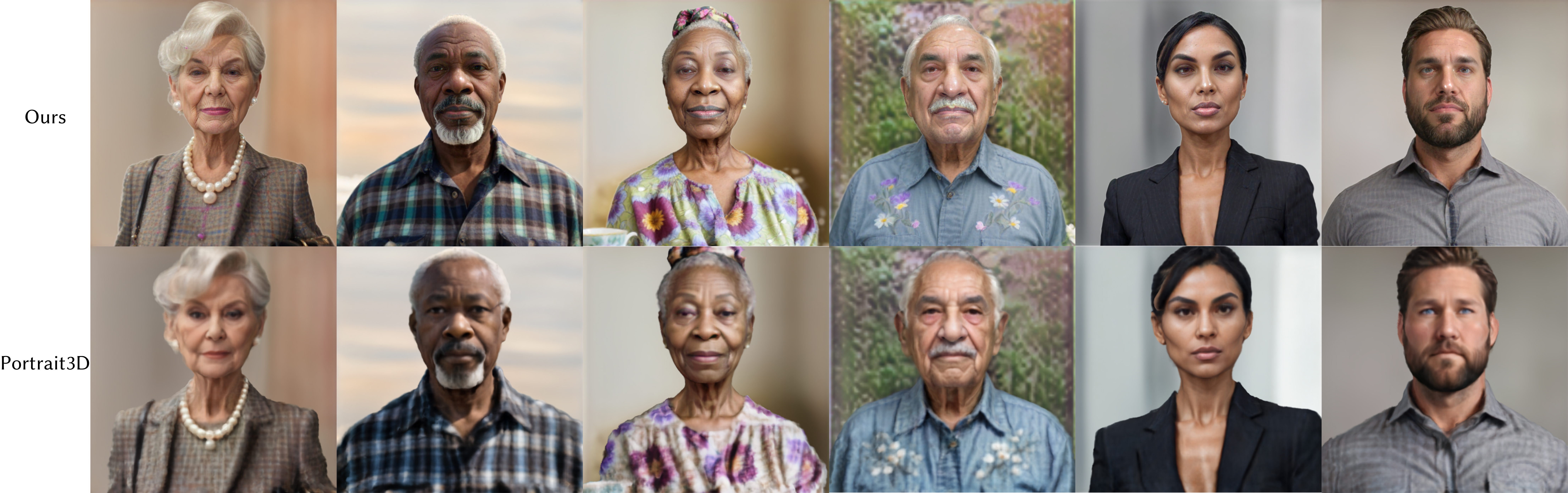}
  
  \caption{ 
      This figure presents frontal renderings comparing our method with Portrait3D. Our results are built upon the outputs of Portrait3D, showcasing the improvements introduced by our approach.
  }
  \label{fig: comparison-with-portrait3d}
\end{figure*}
\section{Comparison with Portrait3D}
\label{sec: Comparison with Portrait3D}
Our method builds upon the outputs of Portrait3D, addressing its limitations in animatability and mitigating the relatively blurry results it produces. In \Cref{fig: comparison-with-portrait3d}, we compare the frontal renderings generated by our approach with those of Portrait3D. Note that we use a different camera setting than Portrait3D, leading to slight variations in the renderings relative to the corresponding Portrait3D results. The background images are generated using Portrait3D's generator. As demonstrated in the figure, our method significantly improves the visual quality of Portrait3D's outputs.

{
Additionally, in \Cref{fig: comparison-with-portrait3d}, we observe strong identity preservation between the Portrait3D initialization and our final avatar for two reasons:
i) 
% In our method, the Portrait3D initialization, the ControlNet, and the diffusion model are conditioned on the same text prompt.
In our method, the Portrait3D initialization, ControlNet, and diffusion model are all conditioned on the same text prompt.
ii) 
% For the majority of the face regions we use small noise levels during the diffusion-guided optimization. As such, most of the avatar’s appearance is preserved, and only high-frequency components are corrected.
For most facial regions, we use low noise levels during the diffusion-guided optimization. As a result, the avatar’s overall appearance is preserved, with only high-frequency components being corrected.
}

% \section{Hair and Clothing Gaussian Splatting}
% \label{sec: Hair and Clothing Gaussian Splatting}
% Since our model explicitly samples hair splats from the hair mesh and clothing splats from the clothing mesh, we can generate hair and clothing splat masks, effectively separating these components from the full 3DGS avatar.

% \Cref{fig: hair_cloth_vis} illustrates the rendering results of the complete avatar alongside its isolated hair and clothing components. Additionally, we demonstrate a simple application by swapping one avatar's hair with that of another.

% Although this separation remains coarse due to the absence of explicit training guidance, we hope it inspires future research to refine and expand upon this capability.
% \begin{figure*}[t]
%   \centering
%   \includegraphics[width=0.99\linewidth]{hair_cloth_vis.pdf}
  
%   \caption{
%       In this figure, we present (a) the full avatar, (b) the isolated hair and clothing components, and (c) the swapping results.
%   }
%   \label{fig: hair_cloth_vis}
% \end{figure*}

\section{Additional Qualitative Comparison}
\label{sec: Additional comparison}
 
In this section, we provide additional qualitative comparisons with HeadStudio \cite{HeadStudio}, TADA \cite{TADA}, HumanGaussian \cite{HumanGaussian}, PortraitGen \cite{PortraitGen}, GPAvatar \cite{GPAvatar}, and GAGAvatar \cite{GAGAvatar}, as shown in \Cref{fig: comparison-2}.
To ensure a fair comparison, motion sequences extracted from the same reference video are used across all methods. For each of the two selected frames, we include results demonstrating camera exploration.

\begin{figure*}[t]
  \centering
  \includegraphics[width=0.7\linewidth]{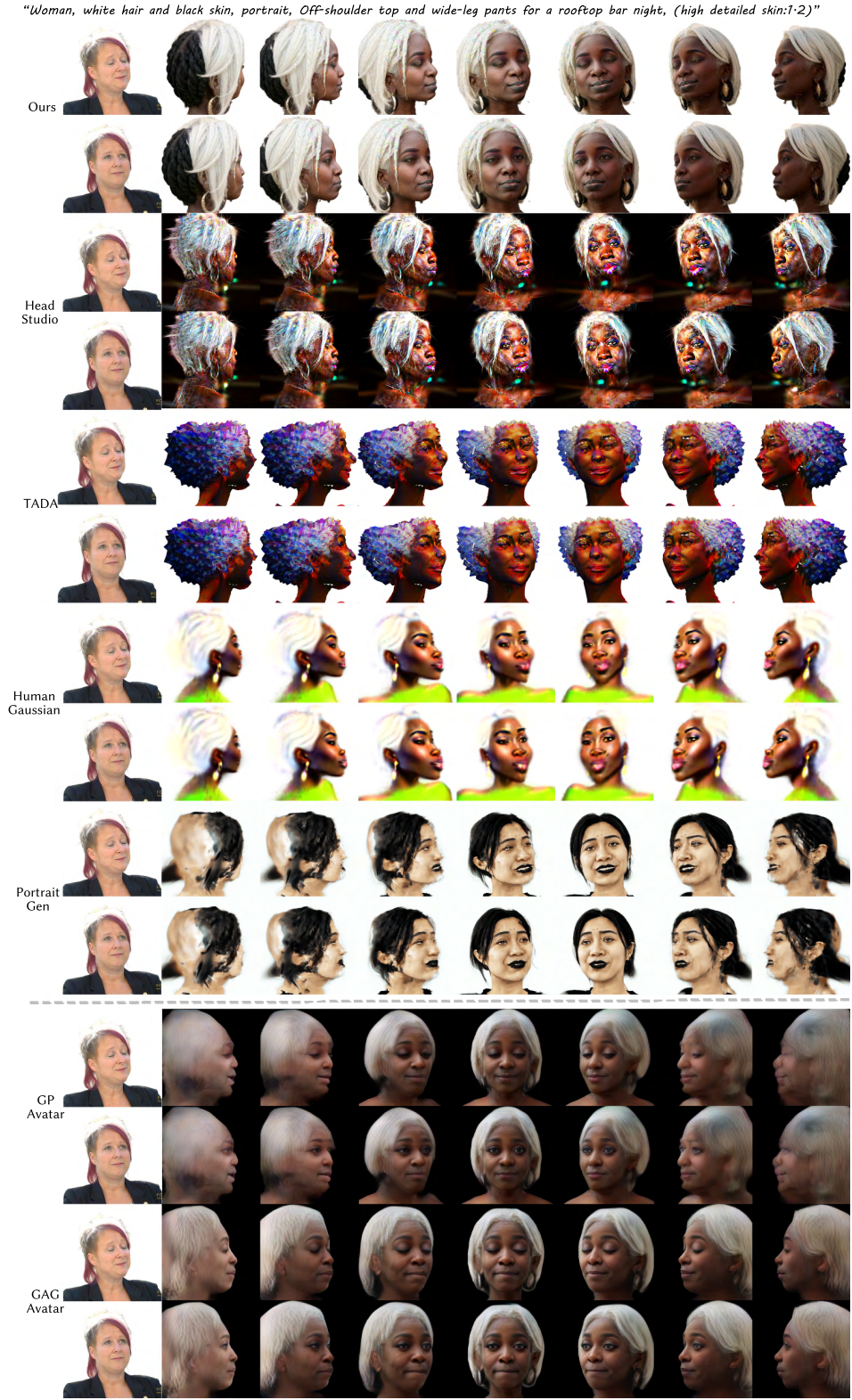}
  
  \caption{
      Comparison with HeadStudio \cite{HeadStudio}, TADA \cite{TADA}, HumanGaussian \cite{HumanGaussian}, PortraitGen \cite{PortraitGen}, GPAvatar \cite{GPAvatar}, and GAGAvatar \cite{GAGAvatar}. 
     While other methods take a text prompt as input (shown at the top), GPAvatar and GAGAvatar use an image as input. 
     The reference images are sourced from the video data in the VFHQ \cite{VFHQ} dataset. 
  }
  \label{fig: comparison-2}
\end{figure*}

\section{Additional Visual Results}
\label{sec: Additional Visual Results} 
Additional results of our method are provided in \Cref{fig: result-1}-\Cref{fig: result-2}. For avatar rendering, we randomly sample expressions and poses from the NeRSemble dataset \cite{NeRSemble}. To demonstrate alignment accuracy, corresponding SMPL-X model renderings are shown alongside the avatars.
Our method is capable of generating 3D animatable avatars that exhibit diversity in gender, age, ethnicities, garments, and hairstyles. These avatars are rendered from various camera views, including the challenging back-view.
\begin{figure*}[t]
  \centering
  \includegraphics[width=0.93\linewidth]{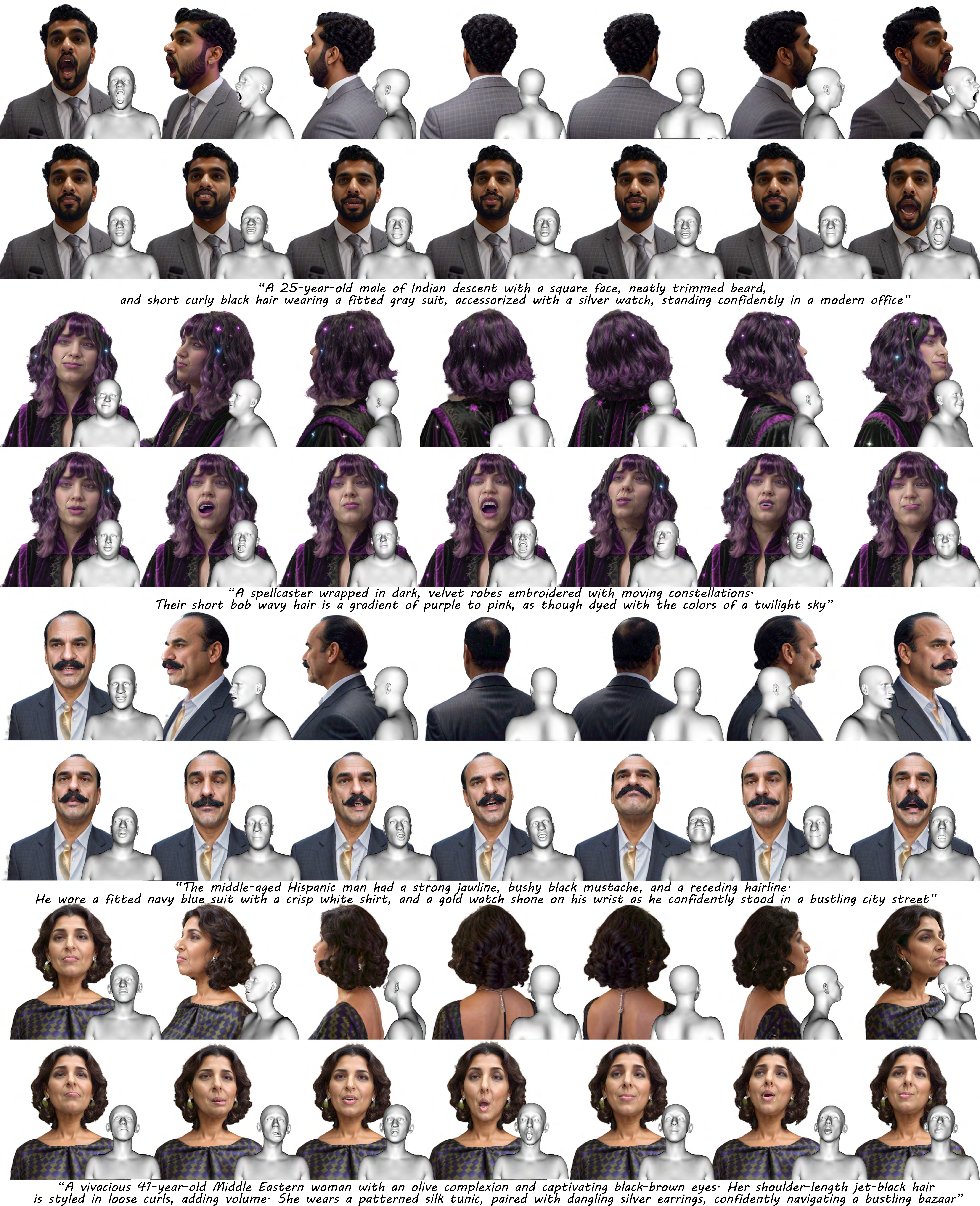}
  
  \caption{ 
     Generated results of our method. For each 3D avatar, the first row presents rendered images with a consistent random expression across different camera views, the second row displays frontal frames with varying expressions and poses, and the corresponding mesh for each avatar is shown at the lower right corner of each rendered image.
  }
  \label{fig: result-1}
\end{figure*}

\begin{figure*}[t]
  \centering
  \includegraphics[width=0.93\linewidth]{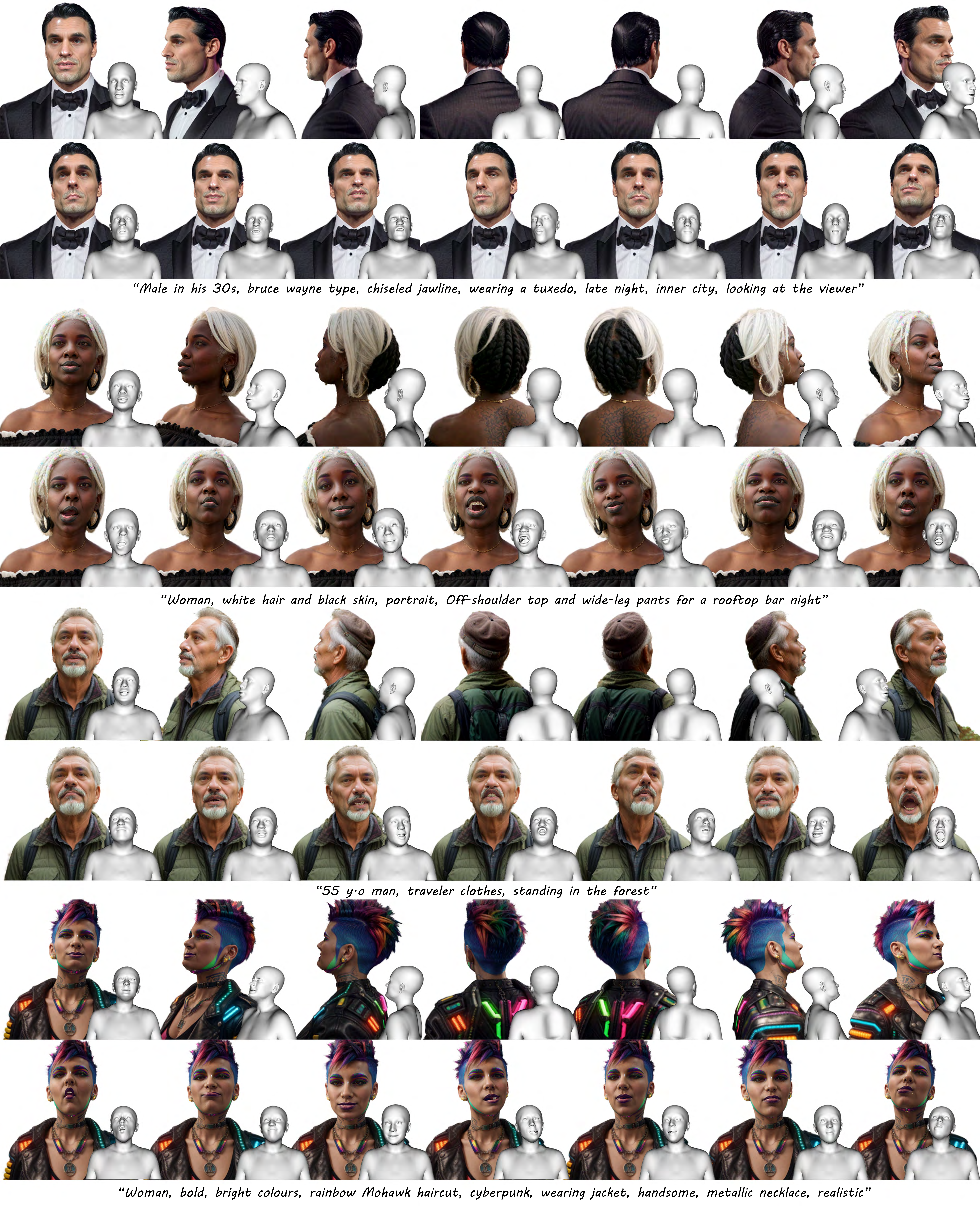}
  
  \caption{ 
     Generated results of our method. For each 3D avatar, the first row presents rendered images with a consistent random expression across different camera views, the second row displays frontal frames with varying expressions and poses, and the corresponding mesh for each avatar is shown at the lower right corner of each rendered image.
  }
  \label{fig: result-2}
\end{figure*}

\section{Discussion}
\label{sec: Discussion}
{While not a technical limitation, our \modelname could be abused for misinformation and fake video generation and may raise ethical concerns. We emphasize the importance of considering the potential implications of generating realistic, animatable 3D avatars. Integrating deepfake detection tools could serve as an effective safeguard. Importantly, applying our method to any specific individual should always require their explicit consent.}

\end{document}